\documentclass[sigconf]{acmart}
\usepackage{algorithm}
\usepackage{algorithmic}

\usepackage{booktabs}
\usepackage{multirow}

\usepackage{amssymb}
\usepackage{svg}
\usepackage{subfig}
\usepackage{tikz}
\usepackage{graphicx}
\usepackage{amsmath}
\usepackage{makecell}
\usepackage{enumitem}
\usepackage{url}
\usepackage[percent]{overpic}
\usepackage{graphicx}
\usepackage{caption}
\usepackage{xcolor}
\usepackage{array}
\usepackage{float}

\makeatletter
\def\@ACM@checkaffil{%
  \if@ACM@instpresent\else
    \ClassWarningNoLine{\@classname}{No institution present for an affiliation}%
  \fi
  \if@ACM@citypresent\else
    \ClassWarningNoLine{\@classname}{No city present for an affiliation}%
  \fi
  \if@ACM@countrypresent\else
    \ClassWarningNoLine{\@classname}{No country present for an affiliation}%
  \fi
}
\makeatother

\copyrightyear{2026}
\acmYear{2026}
\setcopyright{cc}
\setcctype{by}
\acmConference[DAC '26]{63rd ACM/IEEE Design Automation Conference}{July 26--29, 2026}{Long Beach, CA, USA}
\acmBooktitle{63rd ACM/IEEE Design Automation Conference (DAC '26), July 26--29, 2026, Long Beach, CA, USA}
\acmDOI{10.1145/3770743.3803983}
\acmISBN{979-8-4007-2254-7/2026/07}

\begin{document}

\title{The Phantom of PCIe: Constraining Generative Artificial Intelligences for Practical Peripherals Trace Synthesizing}


\author{%
Zhibai Huang\textsuperscript{1},
Chen Chen\textsuperscript{1},
James Yen\textsuperscript{1},
Yihan Shen\textsuperscript{2},
Yongchen Xie\textsuperscript{1},
Zhixiang Wei\textsuperscript{1},
Kailiang Xu\textsuperscript{1},
Yun Wang\textsuperscript{1},
Fangxin Liu\textsuperscript{1,*},
Tao Song\textsuperscript{1},
Mingyuan Xia\textsuperscript{3},
Zhengwei Qi\textsuperscript{1,*}
}

\affiliation{%
  \institution{%
    \textsuperscript{1}Shanghai Jiao Tong University, China
    \textsuperscript{2}Chinese Academy of Sciences, China
    \textsuperscript{3}UltraRISC, China\\[2pt]
    \small{Email: \{paynqueller, chenchen825, jamesyen2202002\}@sjtu.edu.cn},
    \small{shenyihan@lsec.cc.ac.cn},
    \small{xyc007@alumni.sjtu.edu.cn},\\
    \small{\{tonywei\_sjtu, xukl2019, yunwang94, liufangxin, songt333\}@sjtu.edu.cn},
    \small{xiamy@ultrarisc.com},
    \small{qizhwei@sjtu.edu.cn}
  }
}

\renewcommand{\shortauthors}{Huang et al.}

\begin{abstract}
Peripheral Component Interconnect Express (PCIe) is the de facto interconnect standard for high-speed peripherals and CPUs. The development of PCIe devices for emerging applications requires realistic Transaction Layer Packet (TLP) traces that accurately simulate device-CPU interactions. While generative AI offers a promising avenue for synthesizing complex TLP sequences, it is prone to a critical challenge inherent in all generation tasks: hallucination. Naively applying these models often produces traces that violate fundamental PCIe protocol rules, such as ordering and causality, rendering them unusable for device simulation. To resolve this, our work introduces a methodology to bridge the gap between generative AI and high-fidelity device simulation. This paper presents Phantom, a framework that systematically addresses AI-generated hallucinations in TLP synthesis. Phantom achieves this by coupling a generative backbone with a novel post-processing filter that enforces PCIe-specific constraints, effectively eliminating invalid TLP sequences. We validate Phantom's effectiveness by synthesizing TLP traces for an actual PCIe network interface card. Experimental results show that Phantom produces practical, large-scale TLP traces, significantly outperforming existing models, with improvements of up to 1000$\times$ in task-specific metrics and up to 2.19$\times$ in Fréchet Inception Distance (FID) compared to backbone-only methods. The prototype implementation has been made open-source.
\end{abstract}




\maketitle

\section{Introduction}
\label{sec:introduction}

PCIe is the most common interconnection network for peripherals and hosts and is essential for devices like graphics cards, SSDs, and network interface cards. In the PCIe standard, Transaction Layer Packets (TLPs) are the smallest units of information transmitted between PCIe devices \cite{noauthor_pci_nodate}. Understanding a device's TLP transaction patterns from detailed, driver-level traces is crucial for modeling its CPU interaction dynamics and guiding architectural design \cite{kuga2020nettlp,10.1145/3230543.3230560}. Crucially, these patterns can be synthesized into replay-capable traces for in-house device validation and simulation.

\begin{figure}[t]
    \centering
    \includegraphics[width=0.95\linewidth]{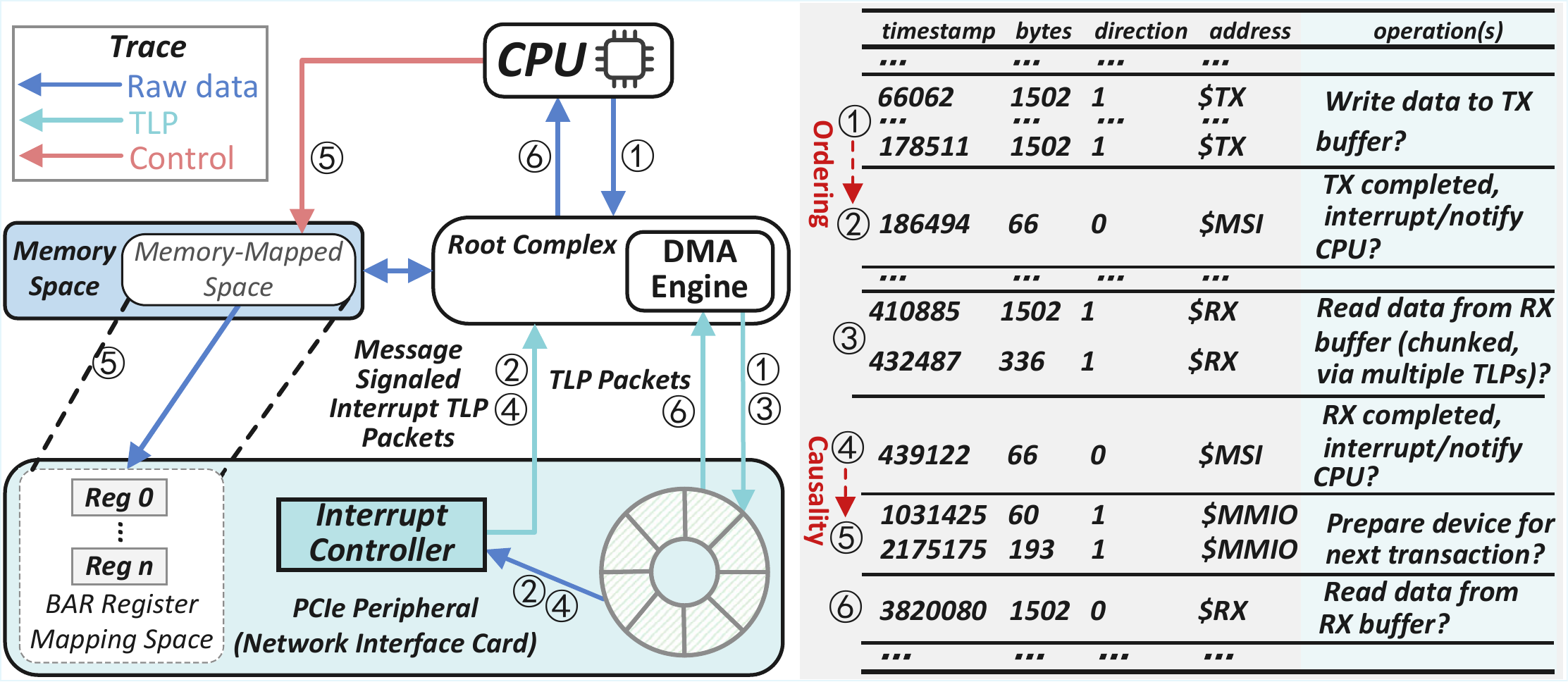}
    \vspace{-1em}
    \caption{Topology of PCIe Devices in Modern Computing Systems. This diagram models a PCIe network interface card, highlighting key concepts like Transaction Layer Packet (TLP), Memory-Mapped Input/Output (MMIO), Direct Memory Access (DMA), Message Signaled Interrupt (MSI), and the transmit (TX) and receive (RX) pathways. Text-based traces are also included to detail the patterns and constraints. 
    }
    \label{fig:pcie_stack}
    \vspace{-2.5em}
\end{figure}

PCIe TLP traces are also complex and restrictive.
Consider Fig.~\ref{fig:pcie_stack}, which illustrates the interaction between a PCIe Network Interface Card (NIC) and the CPU.
NICs enforce critical PCIe transaction constraints: Order (e.g., DMA engine enablement after descriptor ring configuration), causality (interrupt status clearance before new IRQ generation), and size limits (128B-aligned descriptors; 256B TLP payloads). Violating these during trace synthesis corrupts NIC state transitions and renders replicated traffic patterns non-functional. These rigid constraints are precisely where naive generative models fail, as they struggle to learn such deterministic, long-range dependencies, leading to protocol-violating hallucinations.

\begin{figure}[t!]
    \centering
    \setlength{\abovecaptionskip}{0.1cm}
    \setlength{\belowcaptionskip}{-0.5cm}
    \includegraphics[width=0.87\linewidth]{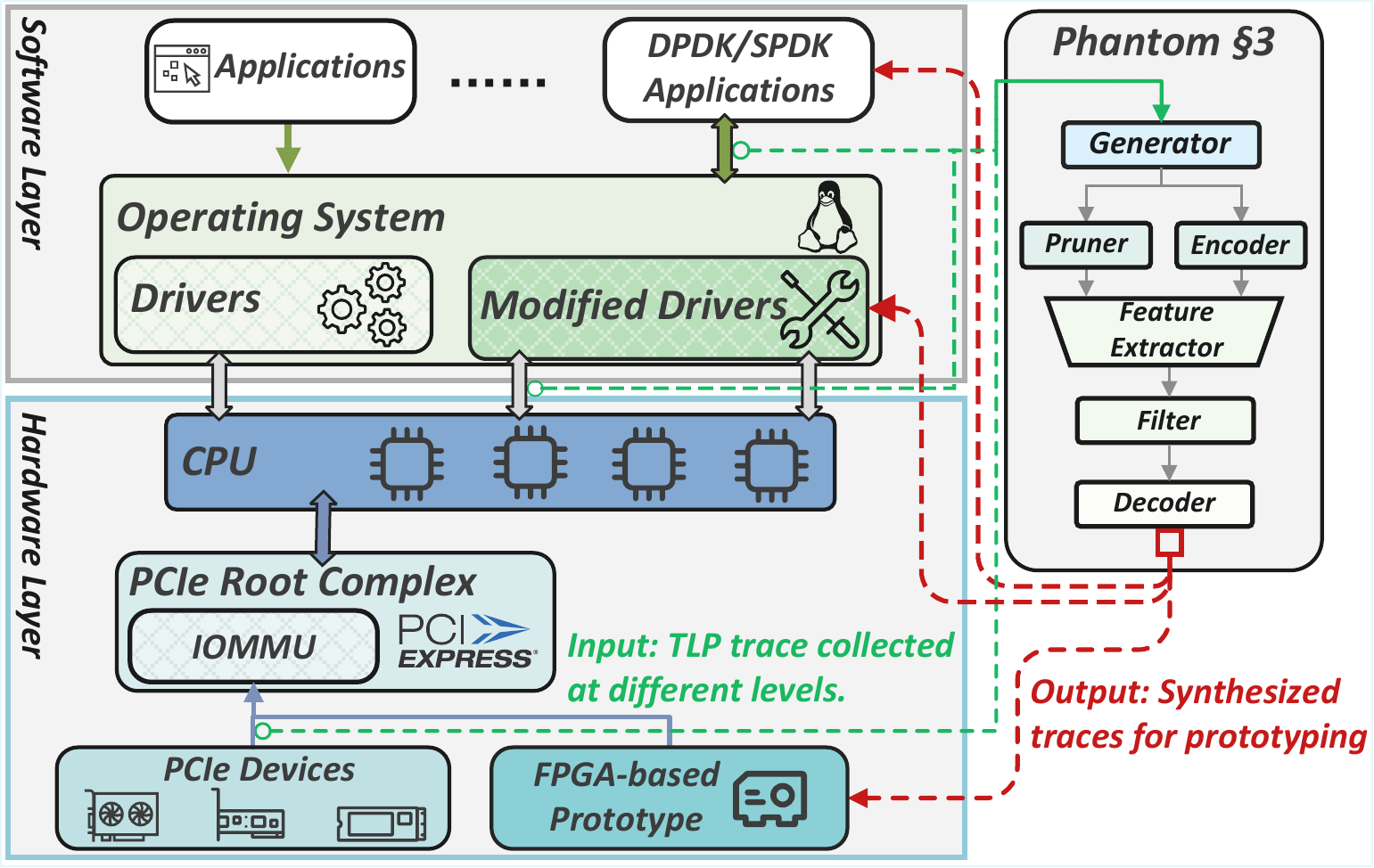}
    \caption{Diagram of Phantom on Target Machine. Phantom receives TLP traffic traces collected from software/hardware layer, and outputs synthesized TLP traffic/packet traces.}
    \label{fig:overview}
    \vspace{-0.5em}
\end{figure}

Traditional statistical-based methods for synthesizing traces often fall short of capturing these complex patterns \cite{ij2018statistics}, and rule-based approaches lack flexibility \cite{thiebaut_synthetic_1992,10.1145/3620666.3651337,maeda2017fast}. Generative AI, assuming sufficient input data, presents a powerful alternative \cite{10.1145/3544216.3544251, NEURIPS2023_604b9fa9}. However, its direct application is hampered by hallucination—the tendency to generate statistically plausible but functionally invalid sequences that violate strict protocol rules, making them useless for high-fidelity device simulation. To bridge the gap between generative AI and device simulation, we introduce Phantom, a framework embodying a novel methodology for constrained TLP synthesis. As illustrated in Fig.~\ref{fig:overview}, Phantom operates independently of the target machine, accepting collected traces to produce synthesized output suitable for replay-based design prototyping.

\begin{figure*}[t]
    \centering
          \setlength{\abovecaptionskip}{0.1cm}
  \setlength{\belowcaptionskip}{-0.4cm}
    \includegraphics[width=0.85\linewidth]{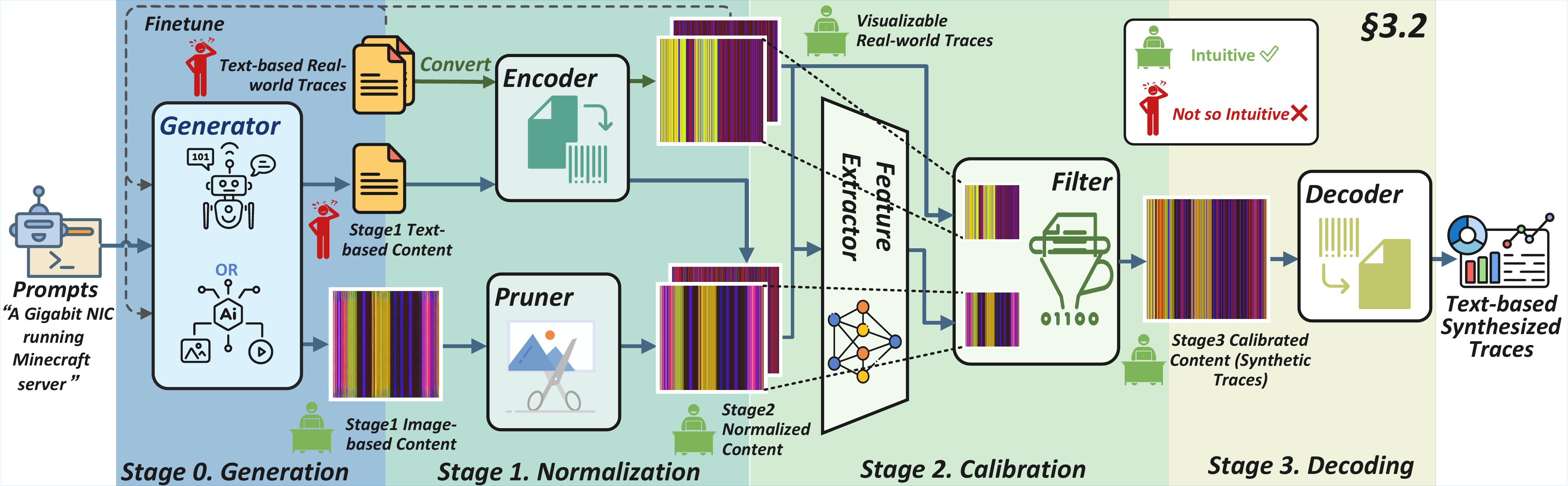}
    \caption{Overview of Phantom with a \textit{1+3} stage pipeline: generation, normalization, calibration, and decoding. Stage 0. The backbone model {\color{red}generates} the initial content. The initial content is {\color{red}normalized} and defects are removed using the TLP trace visualization encoding method. The normalized content is corrected using a dispersion-based calibration method to ensure accuracy and consistency in {\color{red}Calibration} stage. The calibrated content is {\color{red}decoded} to synthesis the final trace.}
    \label{fig:pipeline}
\end{figure*}

The key to Phantom's methodology is a Generate–then–Calibrate mechanism: a generative AI backbone learns the data distribution, followed by a content calibration post-processor that filters hallucinations and enforces protocol compliance. As depicted in Fig.~\ref{fig:pipeline}, this is achieved by first redefining TLP synthesis as an image generation task via a mapping to RGB triplets. A novel convolution-like filter then applies domain expertise to correct the generated trace, effectively removing hallucinated data (later defined as singularities) before a final decoding stage. The key contributions of our work can be summarized as follows:

\textbf{1). }This work is the first to propose a methodology that systematically resolves the hallucination problem in AI-based PCIe trace synthesis. We integrate domain-informed constraints into a post-processing filter, ensuring the generated traces are not only statistically representative but also functionally valid and protocol-compliant.

\textbf{2). }By reformulating PCIe TLP trace synthesis as an image generation problem, we adapt powerful generation models to this domain and enable domain-specific tuning to capture the unique characteristics of PCIe traffic.

\textbf{3). }We present three representative use cases of Phantom, quantitatively demonstrate the advantages of our constrained generative approach over traditional and backbone-only methods, and perform extensive evaluations across diverse generative backbones.

Extensive experiments show that Phantom enhances task-specific metrics by up to $\sim$1,000$\times$ and improves the Fréchet Inception Distance (FID) by up to 2.19$\times$. We have open-sourced the Phantom framework, excluding only the pretrained AI model weights and large language models. The pretrained weights used in our experiments are publicly accessible from their original sources, and the LLMs (such as Llama and GPT4o) are integrated via prompt engineering rather than fine-tuning, ensuring reproducibility of our workflow.

\section{Background}

\subsection{PCIe and Other Interconnects}

The PCIe architecture is layered, primarily consisting of the physical layer (PLP), data link layer (DLLP), and transaction layer (TLP). The physical layer is responsible for the transmission and reception of electrical signals, ensuring the stability of high-speed serial data transmission. The data link layer ensures reliable data transfer by handling error detection and correction, guaranteeing that data packets arrive in the correct order. The transaction layer manages higher-level protocols, such as memory read/write and I/O operations, converting requests into packets.

Due to PCIe's complexity, most other buses (e.g., UART, USB) can be mapped to specific patterns within PCIe TLP traces to some extent. Essentially, all peripheral communications are fundamentally MMIO or PIO interactions, meaning memory access traces can theoretically be obtained at the software layer. This allows the proposed method (in Sec.~\ref{sec:Implementation}) to be applied, albeit with potential variations in effectiveness. This adaptability implies that Phantom requires only minimal adjustments to accommodate trace synthesis for most bus types, making it more generalizable.

\subsection{Trace Synthesis Methods in Related Fields}
Trace synthesis is a well-established area of research, with existing studies exploring various methodologies. These approaches can be broadly classified into three main categories:

\textbf{1). }Simulator-based or Model-fitting Approach: This method replicates the characteristics of the target hardware and software using simulators or mathematical models. Synthetic data is generated through resampling the model \cite{maeda2017fast, thiebaut_synthetic_1992}.

\textbf{2). }Rule-based Approach: Guided by predefined rules, this method utilizes an existing trace as a foundation, performing operations such as "cutting" and "stitching" the data together before adjusting it to produce the final synthesized data \cite{10.1145/3620666.3651337}.

\textbf{3). }Generative AI Approach: This approach leverages generative AI models, which are trained on existing data and guided by user inputs, to generate the necessary synthetic data \cite{10.1145/3544216.3544251, jiang2024netdiffusion}.

Currently, there is no dedicated work focused on PCIe device traces. Existing methods cannot be directly applied, as neglecting the strict constraints of PCIe devices leads to AI hallucination that produces severely distorted and functionally worthless traces.

\section{Design and Implementation}
\label{sec:Implementation}

\subsection{Target Modeling}
PCIe Transaction Layer Packet (TLP) traces are central to system debugging and performance optimization, as they capture transaction types, addresses, payload sizes, directions, and timing. While high-fidelity collection often relies on specialized hardware, this work focuses on synthesizing and analyzing stream-level PCIe TLP traffic at the software layer.

Unlike conventional generative tasks, trace synthesis must strictly respect protocol semantics and system-level causality. 
We reframe trace generation as a spatiotemporal pattern synthesis problem: TLP sequences are mapped to 2D images where pixels encode per-transaction attributes (e.g., access size and direction), ordering becomes spatial layout, and protocol/causal relationships emerge as spatial motifs and clusters. This visual abstraction enables both human interpretability and the use of mature image-generation/ processing backbones for controllable, constraint-aware synthesis.

\subsection{Design of Phantom}
Phantom pipelinely produces raw content with a generic backbone and then calibrates it against real traces to enforce protocol-consistent patterns. The design integrates \textbf{lossless visual representation for TLP traces} and \textbf{dispersion-based calibration}.

\subsubsection{Visual representation (en-/de-coding).}
\begin{figure}[!t]
    \centering
      \setlength{\abovecaptionskip}{0.1cm}
    \includegraphics[width=0.85\linewidth]{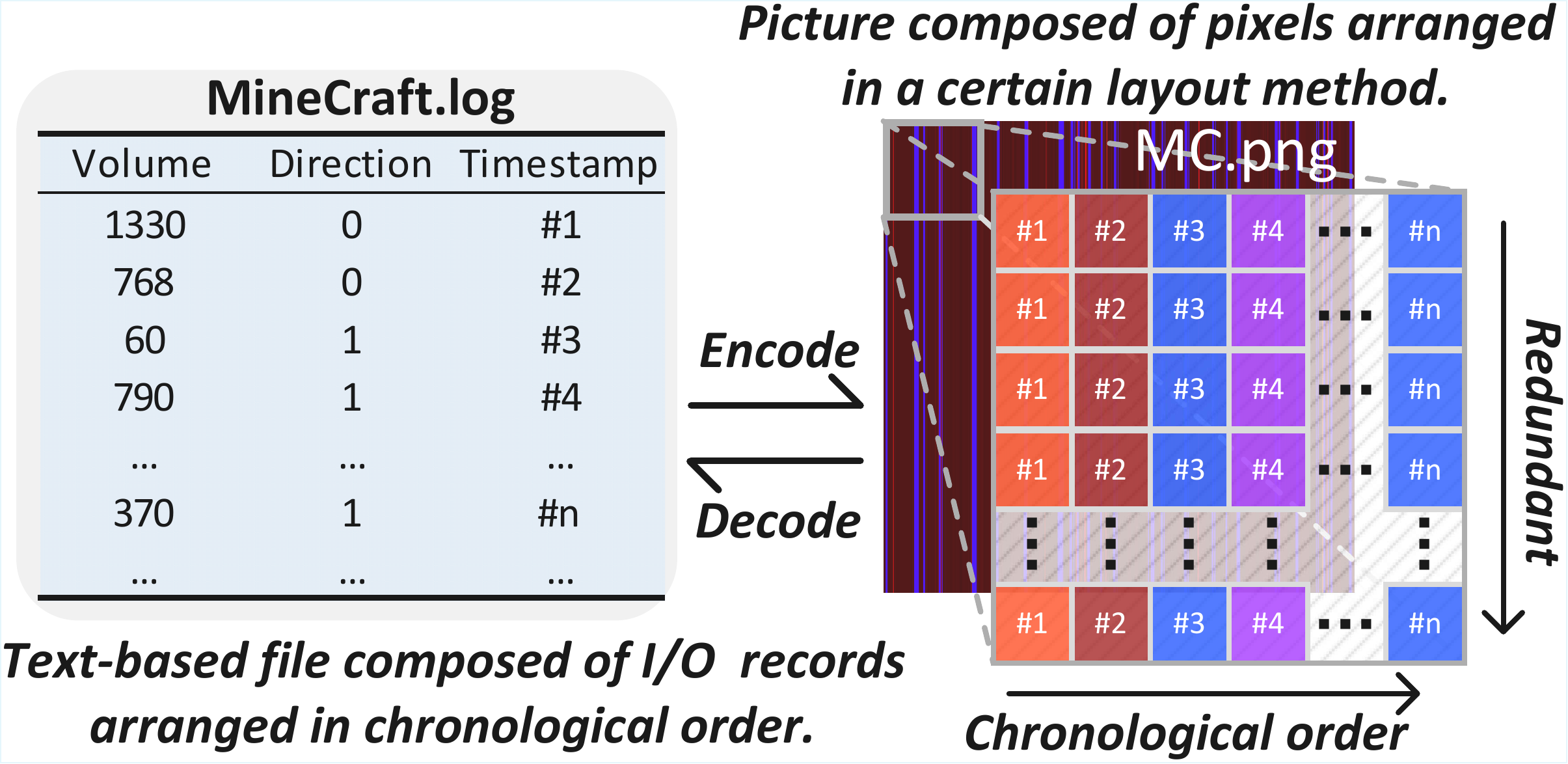}
    \vspace{-0.5em}
    \caption{En-/De-coding Mechanism for TLP Traces.}
    \label{fig:mapping}
    \vspace{-1.5em}
\end{figure}

We encode each TLP record as a pixel and arrange pixels to preserve temporal order. Let a record be
$S_i = (S_i^{\text{byte}}, S_i^{\text{dir}}, S_i^{\text{time}})$, where $S_i^{\text{byte}}$ is the byte size and $S_i^{\text{dir}}\!\in\!\{0,1\}$ denotes direction. Since real traces are time-sorted, we reuse the natural order and drop explicit timestamps; the position serves as a logical time index. A bidirectional mapping to RGB is
\begingroup
\footnotesize
\setlength{\abovedisplayskip}{2pt}
\setlength{\belowdisplayskip}{2pt}
\setlength{\abovedisplayshortskip}{3pt}
\setlength{\belowdisplayshortskip}{3pt}
\begin{equation}
\label{eq:encode}
H_i=\lfloor \tfrac{S_i^{\text{byte}}}{256} \rfloor,\quad
D_i=(H_i,\, S_i^{\text{byte}}-256H_i,\, 255\times S_i^{\text{dir}}).
\end{equation}
\endgroup
The inverse of \eqref{eq:encode} is straightforward, yielding lossless conversion between transaction triplets and RGB pixels. We use horizontal pixel position to encode chronological order and apply column-wise redundancy (identical pixels within a column) for robustness. Fig.~\ref{fig:mapping} illustrates the en-/de-coding mechanism.

\subsubsection{Dispersion-based calibration.}
Backbone outputs may contain \emph{singularities}—locally implausible pixels that violate real TLP patterns. Our calibration stage is designed specifically to identify and correct these hallucinated points. Given a generated, normalized image $x$ and a similar real trace $y$ (found by feature matching), we flag such pixels by combining (1) a global similarity term from learned embeddings and (2) a local, pixel-wise dispersion term.

Let $\widehat{x}$ and $\widehat{y}$ be the tensors after Eq.~\eqref{eq:encode}. A feature extractor $\langle\cdot\rangle$ produces embeddings $x^{\text{enc}}=\langle\widehat{x}\rangle$ and $y^{\text{enc}}=\langle\widehat{y}\rangle$; define an embedding distance
$dist^{\text{enc}}=\mathrm{dist}^{\text{enc}}(x^{\text{enc}},y^{\text{enc}})$ (e.g., cosine distance).
For channel-weighted pixel distance, let $\alpha\in\mathbb{R}^3_{>0}$:
\begingroup
\footnotesize
\setlength{\abovedisplayskip}{2pt}
\setlength{\belowdisplayskip}{2pt}
\setlength{\abovedisplayshortskip}{0pt}
\setlength{\belowdisplayshortskip}{0pt}
\begin{equation}
\label{eq:pixel_distance}
    dist^{\text{pix}}(u,v) = \textstyle\sum_{c=1}^3 \alpha_c \,\big|u_c - v_c\big|.
\end{equation}
\endgroup
Over all pixel locations $(l,m)$, define
\begingroup
\footnotesize
\setlength{\abovedisplayskip}{2pt}
\setlength{\belowdisplayskip}{2pt}
\setlength{\abovedisplayshortskip}{0pt}
\setlength{\belowdisplayshortskip}{0pt}
\begin{equation}
\textstyle
M=\max_{l,m} dist^{\text{pix}}(\widehat{x}^{l,m,:},\widehat{y}^{l,m,:}),\quad
m=\min_{l,m} dist^{\text{pix}}(\widehat{x}^{l,m,:},\widehat{y}^{l,m,:}).
\end{equation}
\endgroup
Let the $n$-neighborhood be
$B_n(l,m)=\{(l_0,m_0)\mid |l-l_0|\le n,\,|m-m_0|\le n\}$ and $\beta\in\mathbb{R}^3_{>0}$ a local weighting. We define the \emph{dispersion score} at $(l,m)$ by
\begingroup
\footnotesize
\setlength{\abovedisplayskip}{0pt}
\setlength{\belowdisplayskip}{0pt}
\setlength{\abovedisplayshortskip}{0pt}
\setlength{\belowdisplayshortskip}{0pt}
\begin{equation}
\label{eq:dispersion}
\mathrm{sgl}^{l,m} =
\frac{dist^{\text{enc}} - m}{M - m}\cdot
\frac{dist^{\text{pix}}(\widehat{x}^{l,m,:},\widehat{y}^{l,m,:})}
{\sum\limits_{(l_0,m_0)\in B_2(l,m)} \sum_{c=1}^3 \beta_c\, \widehat{x}^{\,l_0,m_0,c}}.
\end{equation}
\endgroup

\begin{algorithm}[H]
\caption{Dispersion-Based Trace Calibration (DBTC)}
\label{alg:calibrate_trace}
\footnotesize
\textbf{Input}: normalized generated image $x$, matched real image $y$, threshold $\lambda$, \\ weights $\alpha,\beta$ \\
\textbf{Output}: calibrated image $\mathrm{res}$
\begin{algorithmic}[1]
\STATE compute $dist^{\text{enc}}=\mathrm{dist}^{\text{enc}}(\langle\widehat{x}\rangle,\langle\widehat{y}\rangle)$
\STATE compute per-pixel $d^{l,m}=dist^{\text{pix}}(\widehat{x}^{l,m,:},\widehat{y}^{l,m,:})$ via \eqref{eq:pixel_distance}; get $M=\max d^{l,m}$, $m=\min d^{l,m}$
\FOR{each pixel $(l,m)$}
    \STATE $\mathrm{sgl}^{l,m}\leftarrow$ Eq.~\eqref{eq:dispersion}
    ,$\ $ $\mathrm{res}^{l,m,:}\leftarrow \begin{cases}
        \widehat{y}^{\,l,m,:}, & \mathrm{sgl}^{l,m}>\lambda\\
        \widehat{x}^{\,l,m,:}, & \text{otherwise}
    \end{cases}$    
\ENDFOR
\STATE \textbf{return} $\mathrm{res}$
\end{algorithmic}
\end{algorithm}
\vspace{-2em}

Intuitively, the first factor gates by global mismatch; the second normalizes local deviation by neighborhood energy so that isolated outliers stand out while uniformly large deviations (pattern shift) are not over-penalized. Pixels with $\mathrm{sgl}^{l,m}>\lambda$ (threshold) are replaced by the corresponding pixels from $y$.

\begin{figure*}[ht]
\centering
    \setlength{\abovecaptionskip}{0.1cm}
    \setlength{\belowcaptionskip}{-0.4cm}
\subfloat[Case 1: one pixel’s distance spikes while neighbors are small $\Rightarrow$ large dispersion $\rightarrow$ singularity.]{\label{fig:sgl1}\includegraphics[width=0.3\textwidth]{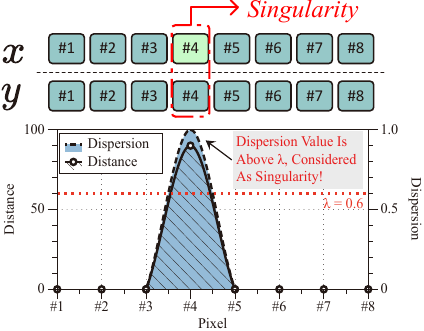}}
\hfill
\subfloat[Case 2: distances are large but with small variation $\Rightarrow$ small dispersion $\rightarrow$ no singularity.]{\label{fig:sgl2}\includegraphics[width=0.3\textwidth]{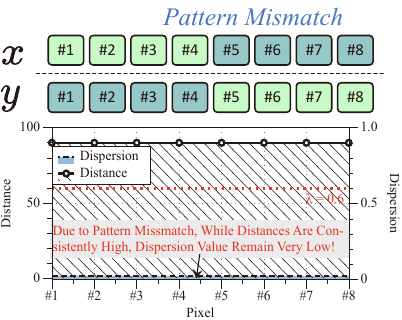}}
\hfill
\subfloat[Case 3: several pixels deviate much more than neighbors $\Rightarrow$ large dispersion $\rightarrow$ singularities.]{\label{fig:sgl3}\includegraphics[width=0.3\textwidth]{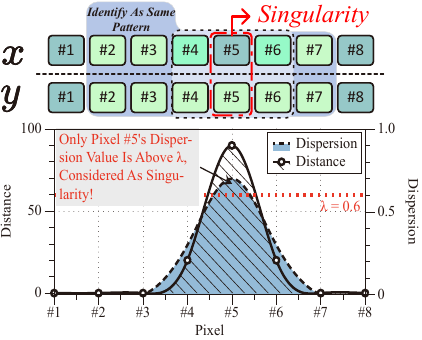}}
\caption{Calibration intuition with generated image $\mathbf{x}$ and real image $\mathbf{y}$. Hue encodes direction; intensity encodes byte size.}

\label{fig:calibration_method}
\end{figure*}

\subsubsection{Pipeline stages.}
\textbf{1) Generation.} A backbone model (text or image) produces raw content with minimal task-specific tuning.

\noindent\textbf{2) Normalization.} Text is encoded via Eq.~\eqref{eq:encode}. Images are pruned, denoised, and normalized to $\widehat{x}$. 

\noindent\textbf{3) Calibration.} We embed the normalized output $\widehat{x}$ and a real-trace corpus to find the nearest real sample $y$ (e.g., using cosine distance or PSNR). We then apply DBTC (Alg.~\ref{alg:calibrate_trace}) for calibration, where $\lambda$ controls the replacement strength from none ($\lambda\!=\!1$) to full ($\lambda\!=\!0$). 
As illustrated in Fig.~\ref{fig:calibration_method}, calibration corrects only pixel dispersions that are significant outliers. Correction is applied when a single pair's dispersion is large (\textbf{Case 1}) or is significantly larger than its neighbors' (\textbf{Case 3}). If all dispersions are uniformly high, no single outlier is identified and no correction occurs (\textbf{Case 2}).

\noindent\textbf{4) Decoding.} The calibrated image is decoded back into a TLP sequence via the inverse of Eq.~\eqref{eq:encode}, ready for downstream validation and design tasks.

This integrated pipeline allows user intervention at any stage and preserves strict round-trip convertibility between traces and images, maximizing control over the backbone model.

\subsubsection{Overheads.}
\noindent\textbf{Runtime Overheads.} The sole runtime cost is the tracing mechanism. By using a small set of representative traces and generative synthesis, our method avoids continuous high-fidelity data capture, ensuring practicality.

\noindent\textbf{Developing Overheads.} Development requires a one-time adaptation of the backbone model (e.g., via fine-tuning or prompt engineering). This initial cost is amortized by the calibration pipeline, which enforces protocol-consistent patterns while minimizing the need for retraining.

\section{Evaluations}
{\renewcommand{\arraystretch}{0.8}
\begin{table*}[t!]
\setlength{\tabcolsep}{10.5pt} 
\scalebox{0.65}{%
\begin{tabular}{c|c|cc|cccccccccc}
\Xhline{2pt}
\multirow{2}{*}{Type}                                                                & \multirow{2}{*}{Generator} & \multicolumn{1}{l}{}                         &         & \multicolumn{10}{c}{Extractor Methods}                                                                                                                                  \\ \cline{3-14} 
                                                                                     &                            & \multicolumn{1}{l|}{
\makecell{Baseline ($\mathit{PE}$)}}                & Attempt & N+c            & N+p            & M+c            & M+p            & I+c            & I+p            & V+c            & V+p            & G+c            & G+p            \\ \hline
\multirow{3}{*}{Worst Case}                                                          & \multirow{3}{*}{Random}    & \multicolumn{1}{c|}{\multirow{3}{*}{2611.6}} & \multicolumn{1}{r|}{best $\downarrow$}    & \textit{20.5$\times$} & \textit{123$\times$}  & \textit{791$\times$}  & \textit{791$\times$}  & \textit{263$\times$}  & \textit{339$\times$}  & \textit{593$\times$}  & \textit{593$\times$}  & \textit{339$\times$}  & \textit{296$\times$}  \\
                                                                                     &                            & \multicolumn{1}{c|}{}                        & \multicolumn{1}{r|}{worst $\downarrow$}   & \textit{14.2$\times$} & \textit{46.6$\times$} & \textit{263$\times$}  & \textit{169$\times$}  & \textit{70.2$\times$} & \textit{87.6$\times$} & \textit{24.3$\times$} & \textit{24.3$\times$} & \textit{47.3$\times$} & \textit{17$\times$}   \\
                                                                                     &                            & \multicolumn{1}{c|}{}                        & \multicolumn{1}{r|}{avg $\downarrow$}     & \textit{17.6$\times$} & \textit{88.9$\times$} & \textit{428$\times$}  & \textit{372$\times$}  & \textit{137$\times$}  & \textit{204$\times$}  & \textit{356$\times$}  & \textit{356$\times$}  & \textit{188$\times$}  & \textit{166$\times$}  \\ \hline
\multirow{6}{*}{\begin{tabular}[c]{@{}c@{}}Common \\ Image \\ Generators\end{tabular}}  & \multirow{3}{*}{GAN}       & \multicolumn{1}{c|}{\multirow{3}{*}{2351.5}} & \multicolumn{1}{r|}{best $\downarrow$}    & \textit{783$\times$}  & \textit{\checkmark}              & \textit{783$\times$}  & \textit{783$\times$}  & \textit{\checkmark}              & \textit{\checkmark}              & \textit{\checkmark}              & \textit{783$\times$}  & \textit{\checkmark}              & \textit{\checkmark}              \\
                                                                                     &                            & \multicolumn{1}{c|}{}                        & \multicolumn{1}{r|}{worst $\downarrow$}   & \textit{167$\times$}  & \textit{\checkmark}              & \textit{391$\times$}  & \textit{391$\times$}  & \textit{\checkmark}              & \textit{\checkmark}              & \textit{\checkmark}              & \textit{783$\times$}  & \textit{783$\times$}  & \textit{783$\times$}  \\
                                                                                     &                            & \multicolumn{1}{c|}{}                        & \multicolumn{1}{r|}{avg $\downarrow$}     & \textit{517$\times$}  & -              & -              & -              & -              & -              & -              & \textit{783$\times$}  & -              & -              \\ \cline{2-14} 
                                                                                     & \multirow{3}{*}{VAE}       & \multicolumn{1}{c|}{\multirow{3}{*}{695}}    & \multicolumn{1}{r|}{best $\downarrow$}    & \textit{\checkmark}              & \textit{\checkmark}              & \textit{\checkmark}              & \textit{\checkmark}              & \textit{\checkmark}              & \textit{\checkmark}              & \textit{\checkmark}              & \textit{\checkmark}              & \textit{\checkmark}              & \textit{\checkmark}              \\
                                                                                     &                            & \multicolumn{1}{c|}{}                        & \multicolumn{1}{r|}{worst $\downarrow$}   & \textit{\checkmark}              & \textit{\checkmark}              & \textit{315$\times$}  & \textit{315$\times$}  & \textit{\checkmark}              & \textit{\checkmark}              & \textit{\checkmark}              & \textit{\checkmark}              & \textit{\checkmark}              & \textit{\checkmark}              \\
                                                                                     &                            & \multicolumn{1}{c|}{}                        & \multicolumn{1}{r|}{avg $\downarrow$}     & -              & -              & -              & -              & -              & -              & -              & -              & -              & -              \\ \hline
\multirow{3}{*}{\begin{tabular}[c]{@{}c@{}}De-facto \\ Image \\ Generator\end{tabular}} & \multirow{3}{*}{SD}        & \multicolumn{1}{c|}{\multirow{3}{*}{2473.1}} & \multicolumn{1}{r|}{best $\downarrow$}    & \textit{\checkmark}              & \textit{449$\times$}  & \textit{749$\times$}  & \textit{749$\times$}  & \textit{281$\times$}  & \textit{281$\times$}  & \textit{1124$\times$} & \textit{1124$\times$} & \textit{2248$\times$} & \textit{1124$\times$} \\
                                                                                     &                            & \multicolumn{1}{c|}{}                        & \multicolumn{1}{r|}{worst $\downarrow$}   & \textit{160$\times$}  & \textit{37.4$\times$}          & \textit{89.9$\times$} & \textit{60.7$\times$} & \textit{68.1$\times$} & \textit{68.1$\times$} & \textit{28.0$\times$} & \textit{28.0$\times$} & \textit{36.6$\times$} & \textit{18.1$\times$} \\
                                                                                     &                            & \multicolumn{1}{c|}{}                        & \multicolumn{1}{r|}{avg $\downarrow$}     & -              & \textit{226$\times$}  & \textit{259$\times$}  & \textit{259$\times$}  & \textit{175$\times$}  & \textit{173$\times$}  & \textit{330$\times$}  & \textit{337$\times$}  & \textit{648$\times$}  & \textit{359$\times$}  \\ \hline
\multirow{3}{*}{\begin{tabular}[c]{@{}c@{}}Common \\ Text\\ Generator\end{tabular}}     & \multirow{3}{*}{LSTM}      & \multicolumn{1}{c|}{\multirow{3}{*}{5915.8}} & \multicolumn{1}{r|}{best $\downarrow$}    & \textit{\checkmark}     & \textit{\checkmark}     & \textit{\checkmark}     & \textit{\checkmark}     & \textit{\checkmark}     & \textit{\checkmark}     & \textit{\checkmark}     & \textit{\checkmark}     & \textit{\checkmark}     & \textit{\checkmark}     \\
                                                                                     &                            & \multicolumn{1}{c|}{}                        & \multicolumn{1}{r|}{worst $\downarrow$}   & \textit{\checkmark}     & \textit{\checkmark}     & \textit{\checkmark}     & \textit{493$\times$}  & \textit{493$\times$}  & \textit{\checkmark}     & \textit{\checkmark}     & \textit{\checkmark}     & \textit{\checkmark}     & \textit{\checkmark}     \\
                                                                                     &                            & \multicolumn{1}{c|}{}                        & \multicolumn{1}{r|}{avg $\downarrow$}     & -              & -              & -              & -              & -              & -              & -              & -              & -              & -              \\ \hline
\multirow{6}{*}{\begin{tabular}[c]{@{}c@{}}Large \\ Language\\ Model\end{tabular}}      & \multirow{3}{*}{Llama}     & \multicolumn{1}{c|}{\multirow{3}{*}{1843.3}} & \multicolumn{1}{r|}{best $\downarrow$}    & \textit{4.78$\times$} & \textit{6.92$\times$} & \textit{\checkmark}     & \textit{\checkmark}     & \textit{\checkmark}     & \textit{6.92$\times$} & \textit{29.2$\times$} & \textit{29.2$\times$} & \textit{\checkmark}     & \textit{279$\times$}  \\
                                                                                     &                            & \multicolumn{1}{c|}{}                        & \multicolumn{1}{r|}{worst $\downarrow$}   & \textit{3.61$\times$} & \textit{24.2$\times$} & \textit{279$\times$}  & \textit{279$\times$}  & \textit{15.1$\times$} & \textit{4.86$\times$} & \textit{837$\times$}  & \textit{837$\times$}  & \textit{455.0$\times$} & \textit{23.9$\times$} \\
                                                                                     &                            & \multicolumn{1}{c|}{}                        & \multicolumn{1}{r|}{avg $\downarrow$}     & \textit{4.19$\times$} & \textit{15.5$\times$} & -              & -              & -              & \textit{5.89$\times$} & \textit{433$\times$}  & \textit{433$\times$}  & -              & \textit{152$\times$}  \\ \cline{2-14} 
                                                                                     & \multirow{3}{*}{GPT4o}   & \multicolumn{1}{c|}{\multirow{3}{*}{834.3}}  & \multicolumn{1}{r|}{best $\downarrow$}    & \textit{12.9$\times$} & \textit{11.9$\times$} & \textit{12.6$\times$} & \textit{17.6$\times$} & \textit{21.1$\times$} & \textit{21.1$\times$} & \textit{13.9$\times$} & \textit{13.3$\times$} & \textit{12.0$\times$} & \textit{14.3$\times$} \\
                                                                                     &                            & \multicolumn{1}{c|}{}                        & \multicolumn{1}{r|}{worst $\downarrow$}   & \textit{9.98$\times$} & \textit{9.60$\times$} & \textit{12.4$\times$} & \textit{12.4$\times$} & \textit{11.7$\times$} & \textit{10.5$\times$} & \textit{12.0$\times$} & \textit{10.1$\times$} & \textit{10.1$\times$} & \textit{9.85$\times$} \\
                                                                                     &                            & \multicolumn{1}{c|}{}                        & \multicolumn{1}{r|}{avg $\downarrow$}     & \textit{11.1$\times$} & \textit{11.2$\times$} & \textit{12.6$\times$} & \textit{14.1$\times$} & \textit{15.1$\times$} & \textit{14.3$\times$} & \textit{12.7$\times$} & \textit{12.4$\times$} & \textit{11.7$\times$} & \textit{11.7$\times$} \\ \Xhline{2pt}
\end{tabular}
}
\caption{Improvement of Transmission Package Errors ($\mathit{PE}$) for Phantom vs. different generator and feature extractor setups. 
Each setup is tested 4 times and harmonic mean of $\mathit{PE}$s is reported. \textbf{`\checkmark' denotes zero $\mathit{PE}$. `-' denotes that mean is not calculable.}}
\label{tab:transmisson_package_error}
\vspace{-2.75em}

\end{table*}
}

\subsection{Experimental Setup}
We evaluate Phantom on the X86 machine used for dataset collection, setting the pixel distance \(\alpha \in \{1, 100, 10000\}\) (Eq.~\ref{eq:pixel_distance}), dispersion \(\beta \in \left\{\tfrac{1}{12}, \tfrac{1}{6}, \tfrac{1}{2}, \tfrac{1}{6}, \tfrac{1}{12}\right\}\) (Eq.~\ref{eq:dispersion}), and an acceptance threshold ranging from \(1\) down to \(1 \times 10^{-10}\). Phantom integrates a diverse suite of generators, including GAN, VAE, LSTM, and fine-tuned foundation models like Stable Diffusion~\cite{rombach2021highresolution}, Llama~\cite{touvron2023llama}, and GPT4o~\cite{achiam2023gpt}, alongside additional models trained on our \(\sim30{,}000\)-image dataset of browsing, gaming, streaming, VNC, and web crawling workloads. For feature extraction, we compare five distinct models—naive (N), MobileNet (M), InceptionV3 (I), VAE-encoder (V), and GAN-discriminator (G)—using cosine similarity (c) and PSNR (p) as similarity metrics\nocite{Howard2017MobileNetsEC,7780677}.

\subsection{Evaluation Metrics}
To quantify the extent of AI hallucination in this domain, we design two new evaluation metrics (Transmission Package Error and Transmission Traffic Error). These metrics are tailored to assess the protocol rationality and correctness of generated TLP transactions, effectively measuring the severity of hallucinations. The definitions of two metrics are as follows:

Transmission Package Error ($\mathit{PE}$) and Transmission Traffic Error ($\mathit{TE}$) are defined as:
\begingroup
\footnotesize
\setlength{\abovedisplayskip}{0pt}
\setlength{\belowdisplayskip}{2pt}
\setlength{\abovedisplayshortskip}{0pt}
\setlength{\belowdisplayshortskip}{0pt}
\setlength{\jot}{1pt}
\begin{align}
\mathit{PE}
&= \sum_{i=1}^{T} \Bigl(
      \textstyle\sum_{j=1}^{m} w_j\bigl( |\Delta_{0,j,i}| + |\Delta_{1,j,i}| \bigr)
      + w_t\Bigl|\textstyle\sum_{j=1}^{m} \Delta_{0,j,i}\Bigr|
      + w_t\Bigl|\textstyle\sum_{j=1}^{m} \Delta_{1,j,i}\Bigr|
    \Bigr), \label{eq:pe} \\
\mathit{TE}
&= \sum_{i=1}^{T} \Bigl(
      \textstyle\sum_{j=1}^{m} \sum_{\text{dir}=0}^{1} w_{j,\text{dir}}\,|\Delta_{\text{dir},j,i}|
      + w_t \textstyle\sum_{j=1}^{m} |\Delta_{0,j,i}| \notag\\
&\quad
      + \textstyle\sum_{\text{dir}=0}^{1} w_{\text{total},\text{dir}}
        \Bigl|\textstyle\sum_{j=1}^{m} \Delta_{\text{dir},j,i}\Bigr|
      + w_t \textstyle\sum_{j=1}^{m} |\Delta_{1,j,i}|
    \Bigr). \label{eq:te}
\end{align}
\endgroup
Where, \(\mathbf{w}_t\) is the penalty weight for total throughput or wrong direction. 
\(\mathbf{w}_j\) is the penalty weight vector for each segment, including \(\mathbf{w}_{j,0}\)/\(\mathbf{w}_{j,1}\) for the send/receive directions respectively. 
\(\mathbf{w}_{\text{total}, \text{dir}}\) is the penalty weight for the total send/receive direction.
\(\text{D}_{\text{synth}, \text{dir}, j, i}\) and \(\text{D}_{\text{real}, \text{dir}, j, i}\) are the synthetic and real data at the \(i\)-th time frame, \(j\)-th segment, and \(\text{dir}\) direction. The difference between synthetic and real data is \(\mathbf{\Delta}_{\text{dir}, j, i} = \text{D}_{\text{synth}, \text{dir}, j, i} - \text{D}_{\text{real}, \text{dir}, j, i}\). \(T\) is the number of time frames, and \(m\) is the number of segments.

We also use Fréchet Inception Distance (FID) \cite{NIPS2017_8a1d6947} for the traditional generative task metric. To calculate $\mathit{PE}$ and $\mathit{TE}$, each generated image is compared to the closest real content, as identified by the feature extractor. A batch of four generated images is processed alongside the real trace dataset to compute the FID score. Transmission errors are then obtained by taking the harmonic average of the four images. We prioritize minimizing $\mathit{TE}$, which is considered the most important of the three metrics.

From a semantic standpoint, the three metrics we adopt each assess the statistical similarity between a synthesized trace and the chosen real benchmark trace along a distinct dimension. The $\mathit{PE}$ and $\mathit{TE}$ quantify discrepancies in traffic dynamics and workload composition, respectively, while the FID measures divergence between the synthesized trace and a set of real traces within a common projection space. Hence, superior performance on these metrics implies—statistically—that the synthesized trace more closely resembles the benchmark trace, i.e., it appears more ``realistic.'' 

\subsection{Example Applications of Phantom}

\subsubsection{Augmentation of Trace Data}
\begin{figure}[t!]
\centering
    \setlength{\abovecaptionskip}{0.1cm}
    \setlength{\belowcaptionskip}{-0.5cm}
\includegraphics[width=0.9\linewidth]{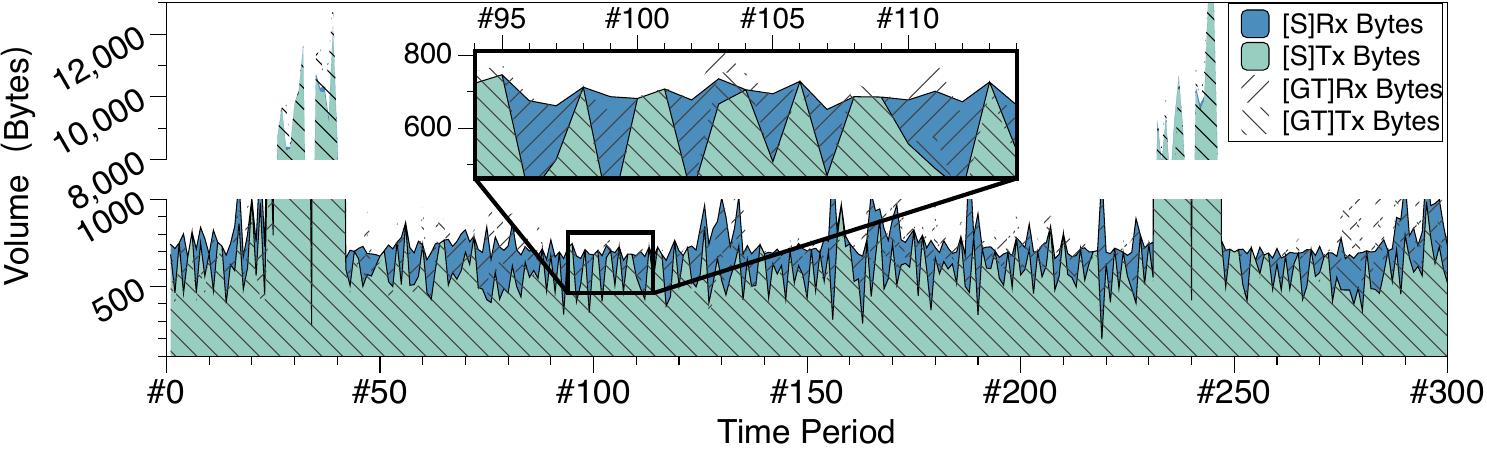}\vspace{-0.5em}
\caption{A Synthesized Trace Example. 
The synthesized trace (S) demonstrates trends in both transmit (Tx) and receive (Rx) traffic that closely match the given ground truth (GT).
}
\label{fig:expand_trace}
\end{figure}

As a trace synthesizer, the most basic use case of Phantom is to rapidly and accurately generate high-quality traces from partial existing data, thereby augmenting the trace dataset. We trained a diffusion-based backbone generator using a subset of the collected data and specified calibration samples to synthesize new trace files. The results, presented in Fig.~\ref{fig:expand_trace}, demonstrate that the synthesized traces exhibit similar overall traffic trends and proportions to the target correction files, though they are not identical. This outcome highlights the feasibility of using Phantom for data augmentation.

\subsubsection{Efficient Fuzzing Test Case Generation}

\begin{figure}[t]
\centering
    \setlength{\abovecaptionskip}{0.1cm}
    \setlength{\belowcaptionskip}{-0.5cm}
\begin{tikzpicture}

\begin{scope}[shift={(0, 0)}]
    \node[anchor=north west, inner sep=0] at (0, 0)
        {\includegraphics[width=1.25cm, height=0.625cm]{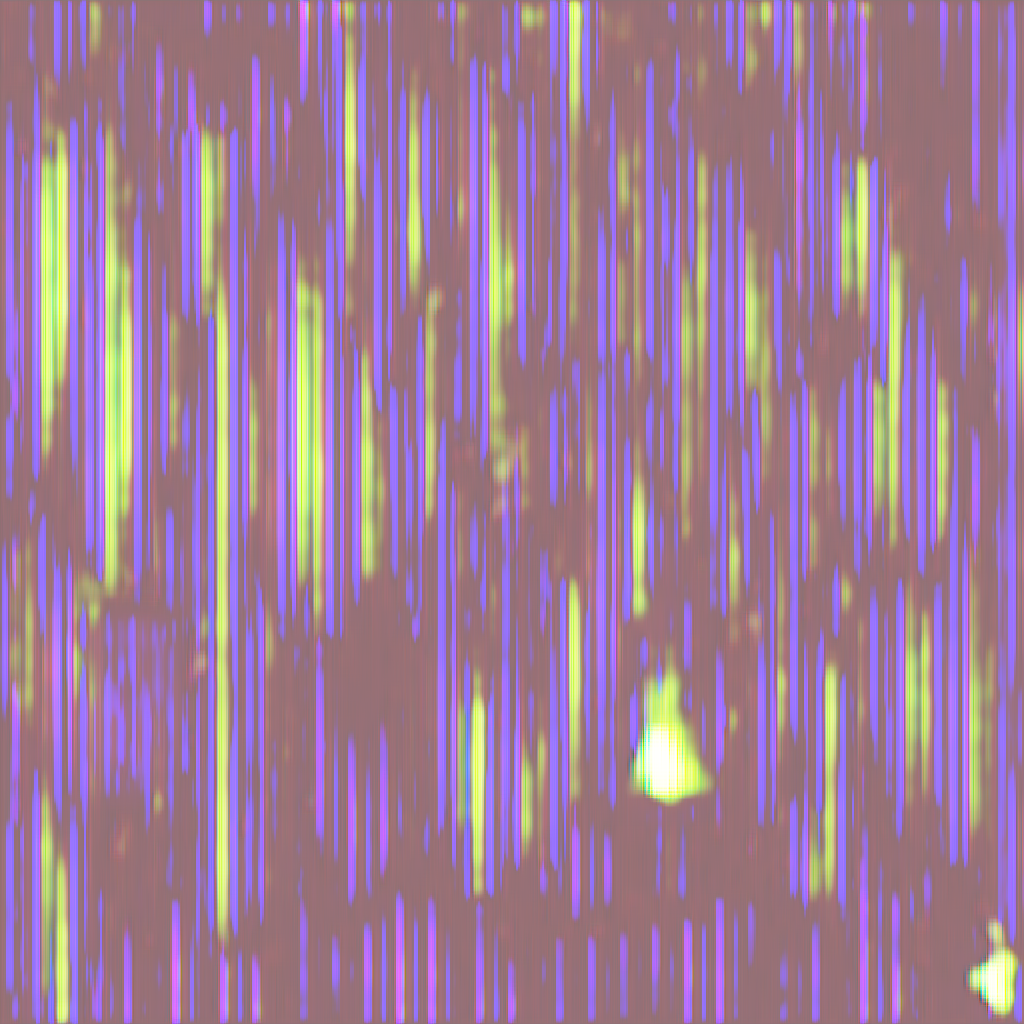}};
    \node[anchor=north west, inner sep=0] at (0, -0.70)
        {\includegraphics[width=1.25cm, height=0.625cm]{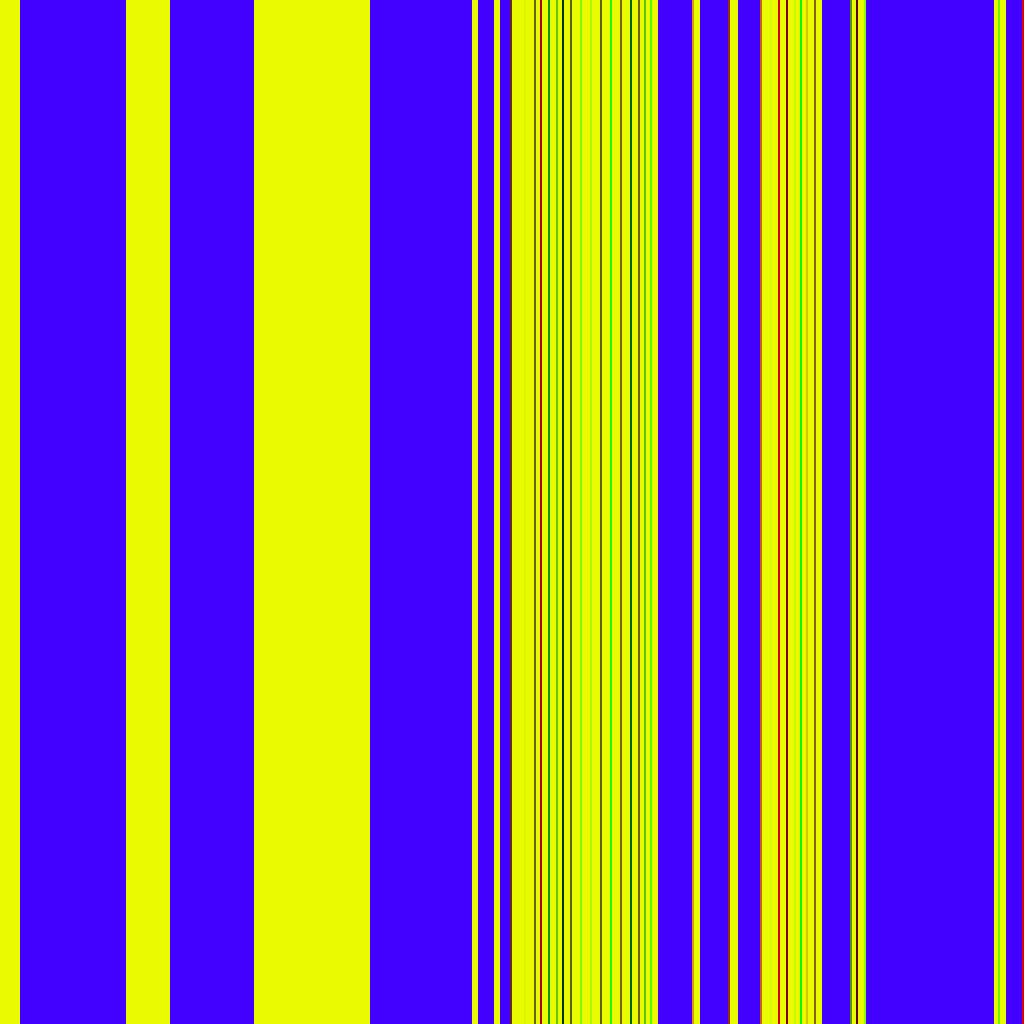}};
    \node[anchor=north west, inner sep=0] at (1.325, 0)
        {\includegraphics[width=2.55cm, height=1.275cm]{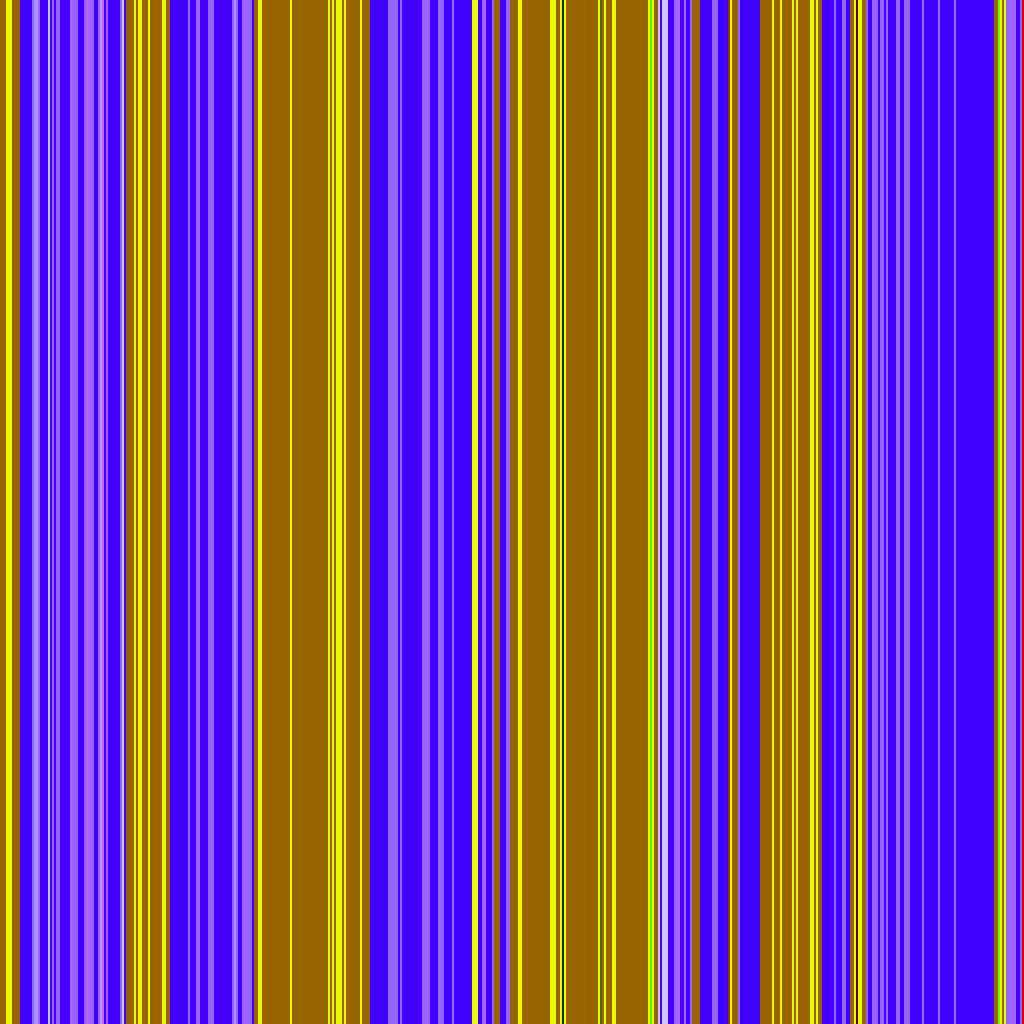}};
    \node[anchor=north west, inner sep=0] at (0, -1.45) {\footnotesize GAN};
    \node[anchor=north west, inner sep=2, text=white] at (0, 0) {\fontsize{9pt}{9pt}\selectfont R};
    \node[anchor=north west, inner sep=2, text=white] at (0, -0.70) {\parbox{1.25cm}{\fontsize{9pt}{9pt}\selectfont PK}};
    \node[anchor=north west, inner sep=2, text=white] at (1.325, 0) {\fontsize{9pt}{9pt}\selectfont P};

    \draw[red, very thick] (0.65, 0.05) rectangle (0.9, -1.35);
    \draw[red, very thick] (1.35+2*0.65, 0.05) rectangle (1.35+2*0.9, -1.35);
\end{scope}

\begin{scope}[shift={(4, 0)}]
    \node[anchor=north west, inner sep=0] at (0, 0)
        {\includegraphics[width=1.25cm, height=0.625cm]{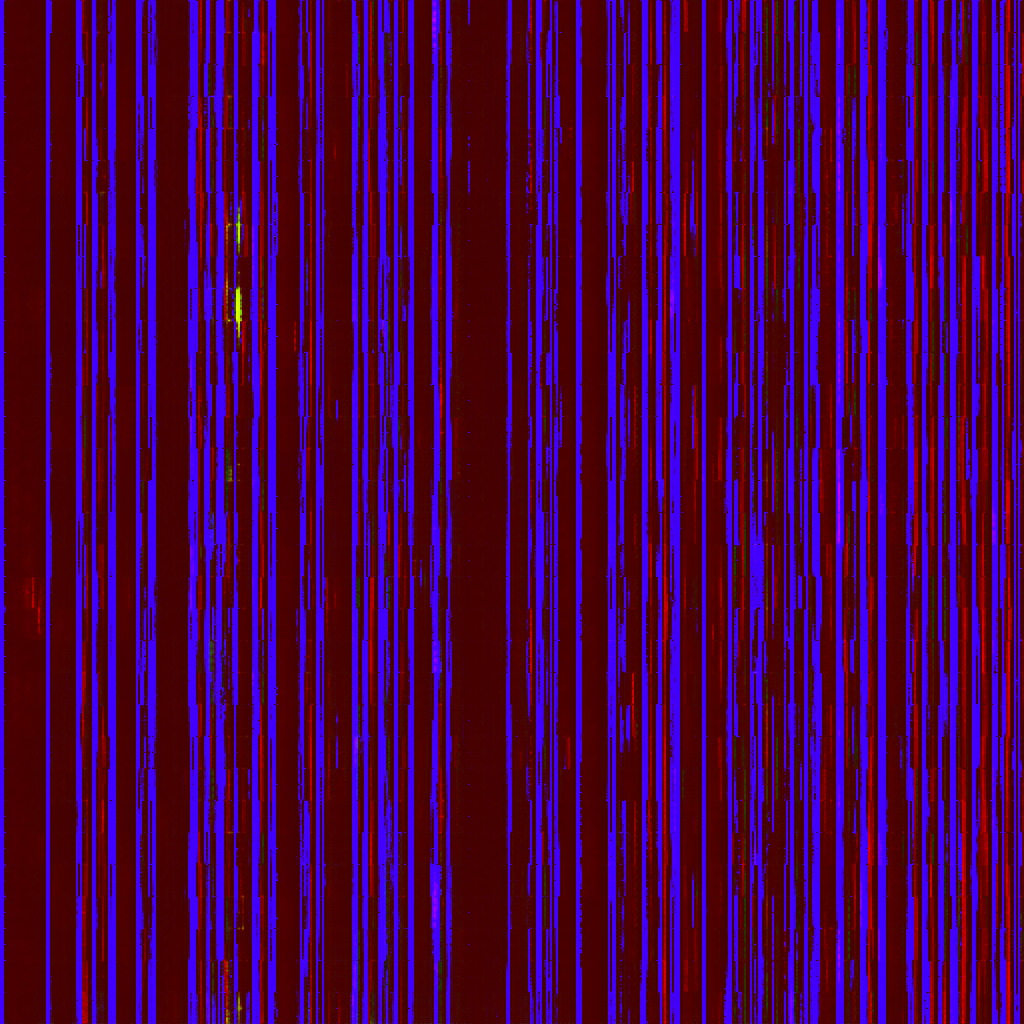}};
    \node[anchor=north west, inner sep=0] at (0, -0.70)
        {\includegraphics[width=1.25cm, height=0.625cm]{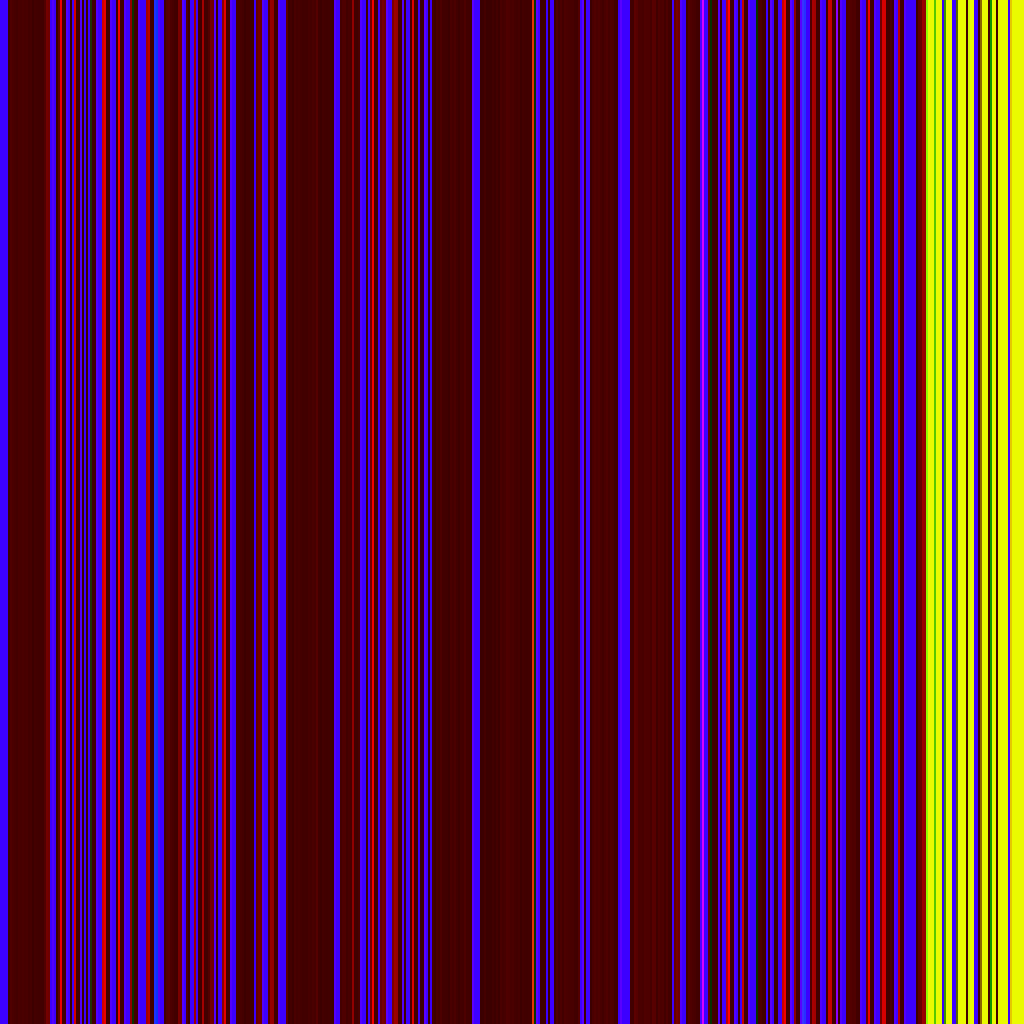}};
    \node[anchor=north west, inner sep=0] at (1.325, 0)
        {\includegraphics[width=2.55cm, height=1.275cm]{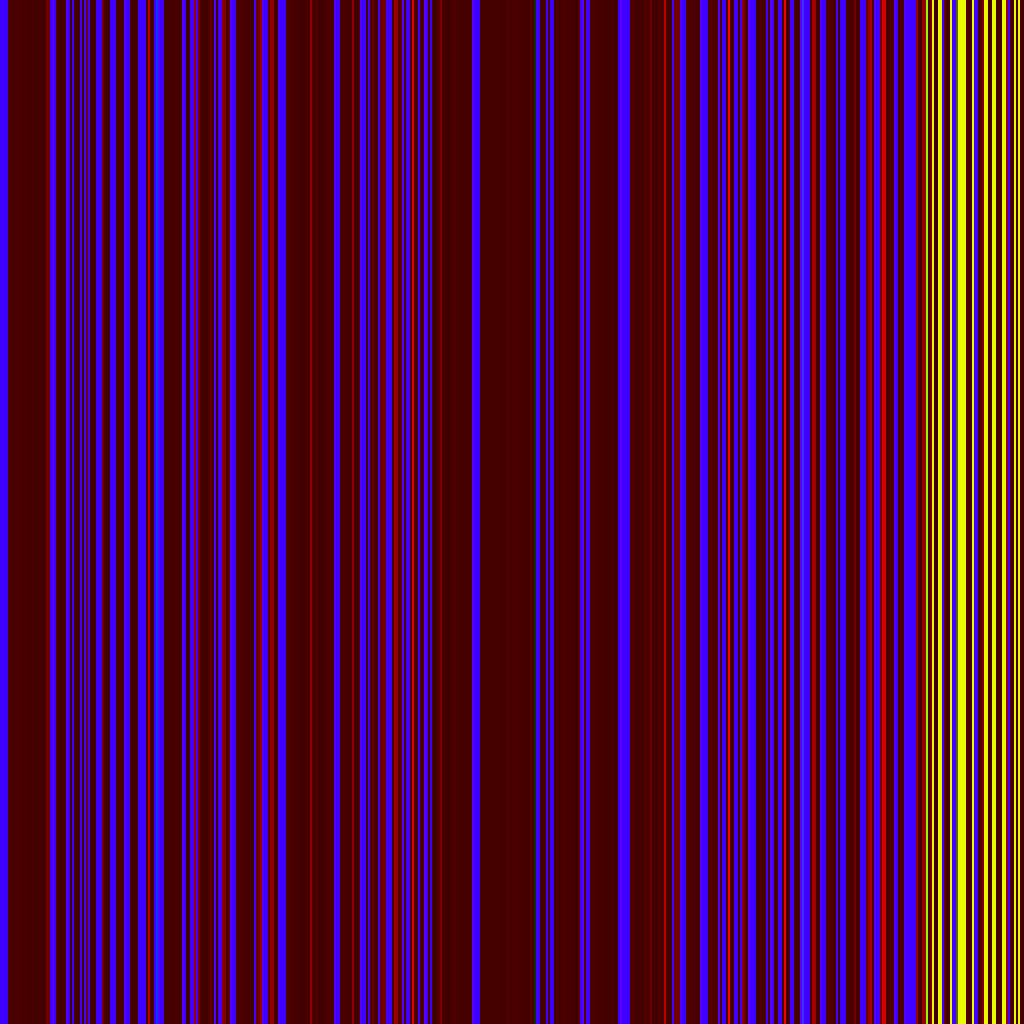}};
    \node[anchor=north west, inner sep=0] at (0, -1.45) {\footnotesize VAE};
    \node[anchor=north west, inner sep=2, text=white] at (0, 0) {\fontsize{9pt}{9pt}\selectfont R};
    \node[anchor=north west, inner sep=2, text=white] at (0, -0.70) {\parbox{1.25cm}{\fontsize{9pt}{9pt}\selectfont PK}};
    \node[anchor=north west, inner sep=2, text=white] at (1.325, 0) {\fontsize{9pt}{9pt}\selectfont P};

    \draw[red, very thick] (1.1, 0.05) rectangle (1.25, -1.35);
    \draw[red, very thick] (1.35+2*1.1, 0.05) rectangle (1.35+2*1.25, -1.35);
\end{scope}

\begin{scope}[shift={(0, -1.75)}]
    \node[anchor=north west, inner sep=0] at (0, 0)
        {\includegraphics[width=1.25cm, height=0.625cm]{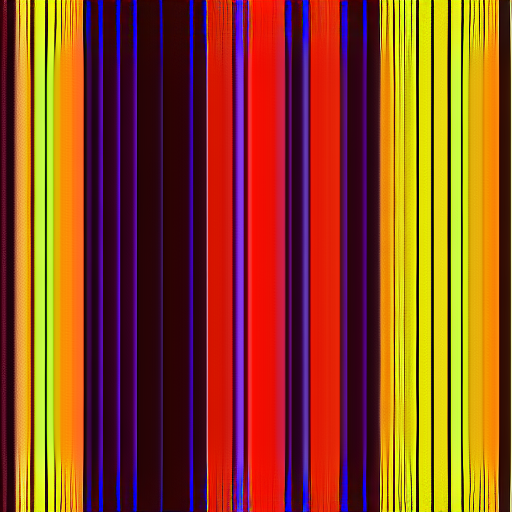}};
    \node[anchor=north west, inner sep=0] at (0, -0.70)
        {\includegraphics[width=1.25cm, height=0.625cm]{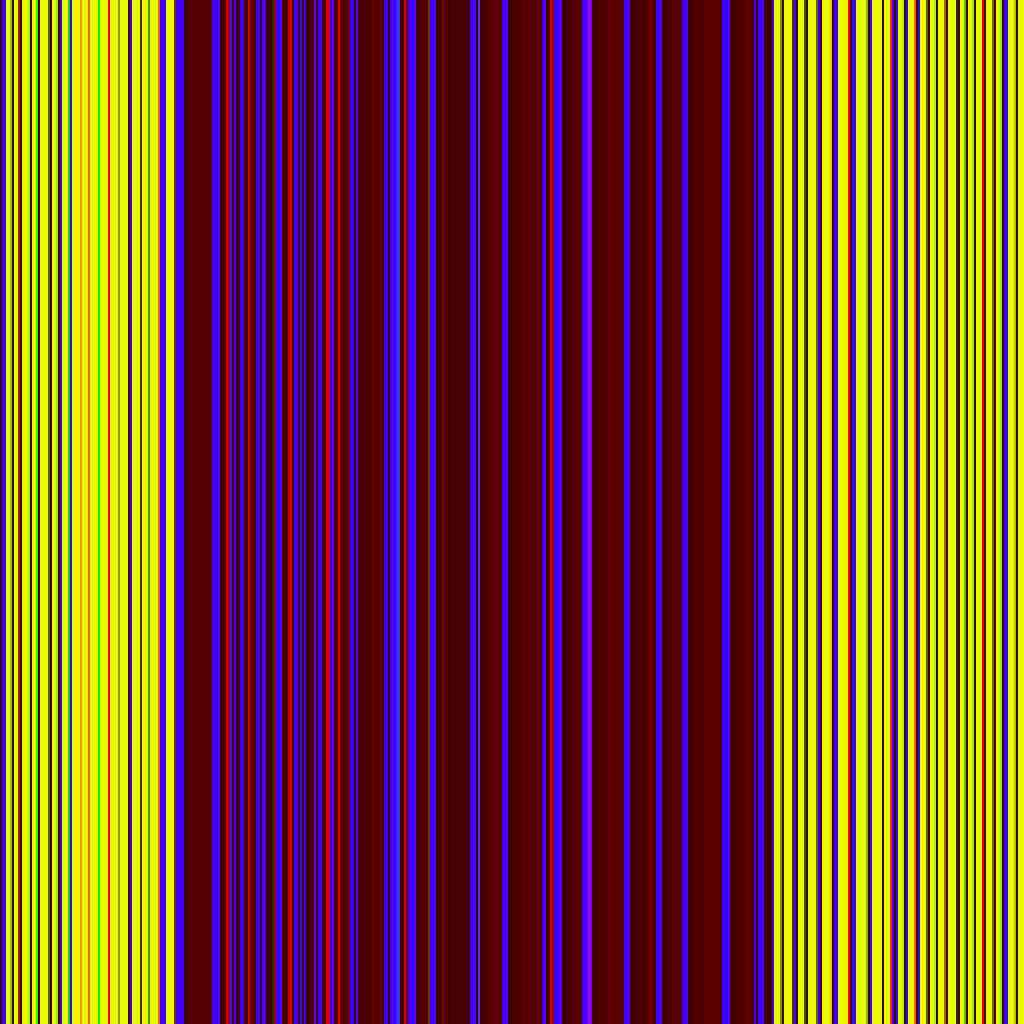}};
    \node[anchor=north west, inner sep=0] at (1.325, 0)
        {\includegraphics[width=2.55cm, height=1.275cm]{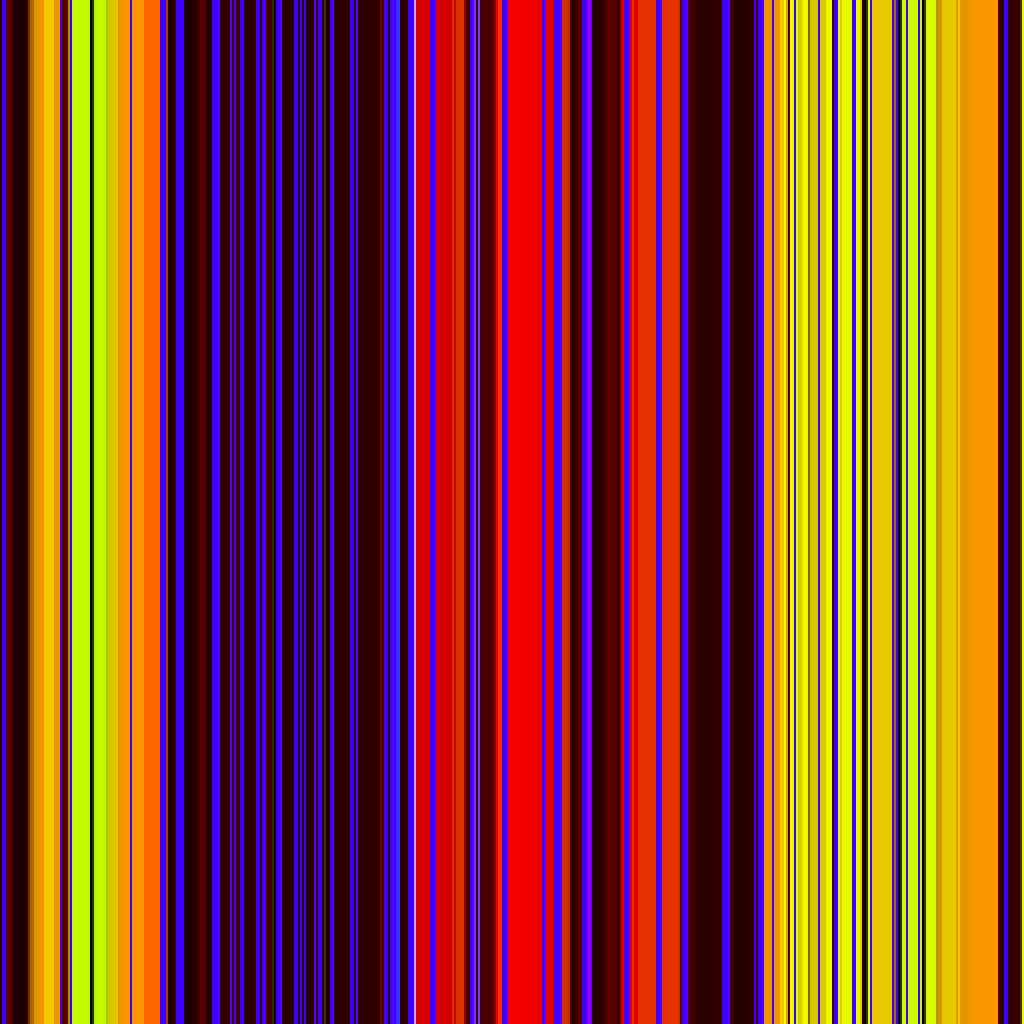}};
    \node[anchor=north west, inner sep=0] at (0, -1.45) {\footnotesize Stable Diffusion};
    \node[anchor=north west, inner sep=2, text=white] at (0, 0) {\fontsize{9pt}{9pt}\selectfont R};
    \node[anchor=north west, inner sep=2, text=white] at (0, -0.70) {\parbox{1.25cm}{\fontsize{9pt}{9pt}\selectfont PK}};
    \node[anchor=north west, inner sep=2, text=white] at (1.325, 0) {\fontsize{9pt}{9pt}\selectfont P};

    \draw[red, very thick] (0.475, 0.05) rectangle (0.875, -1.35);
    \draw[red, very thick] (1.35+2*0.475, 0.05) rectangle (1.35+2*0.875, -1.35);
\end{scope}

\begin{scope}[shift={(4, -1.75)}]
    \node[anchor=north west, inner sep=0] at (0, 0)
        {\includegraphics[width=1.25cm, height=0.625cm]{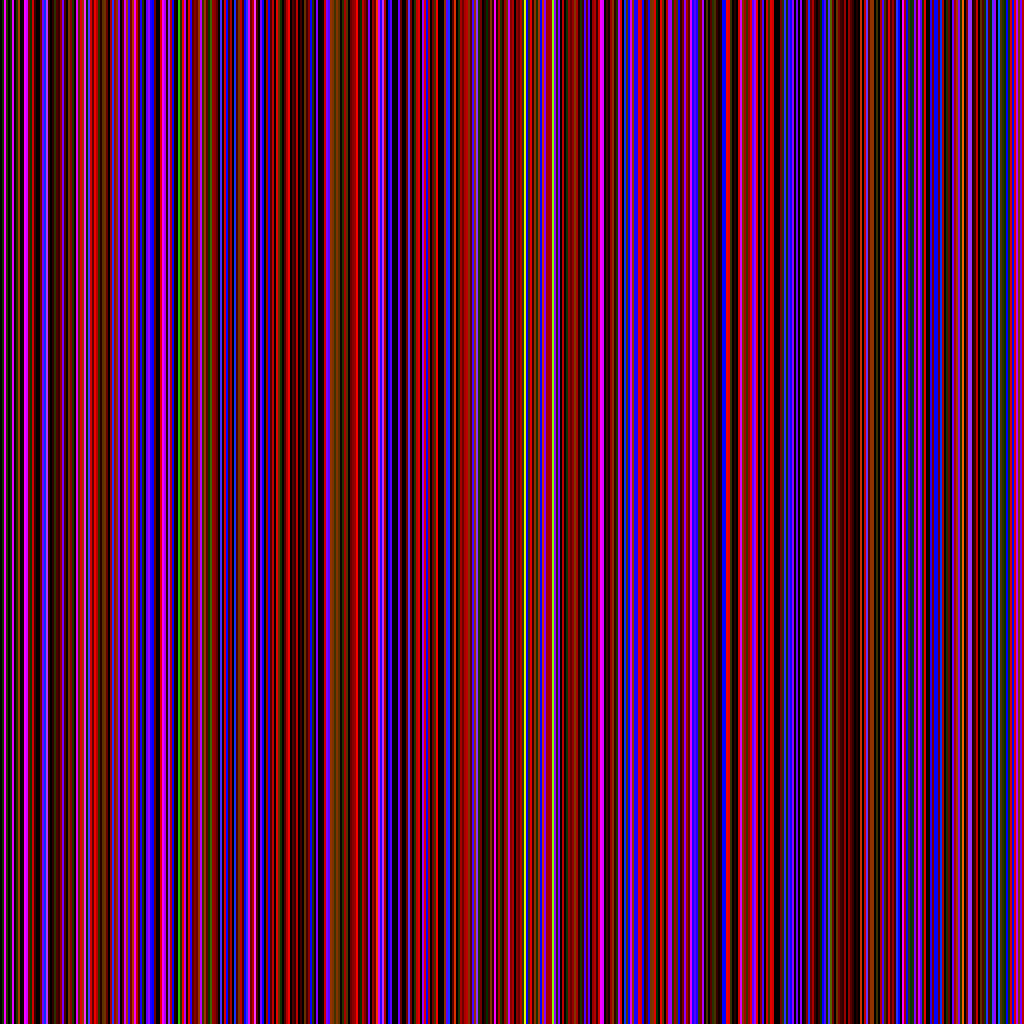}};
    \node[anchor=north west, inner sep=0] at (0, -0.70)
        {\includegraphics[width=1.25cm, height=0.625cm]{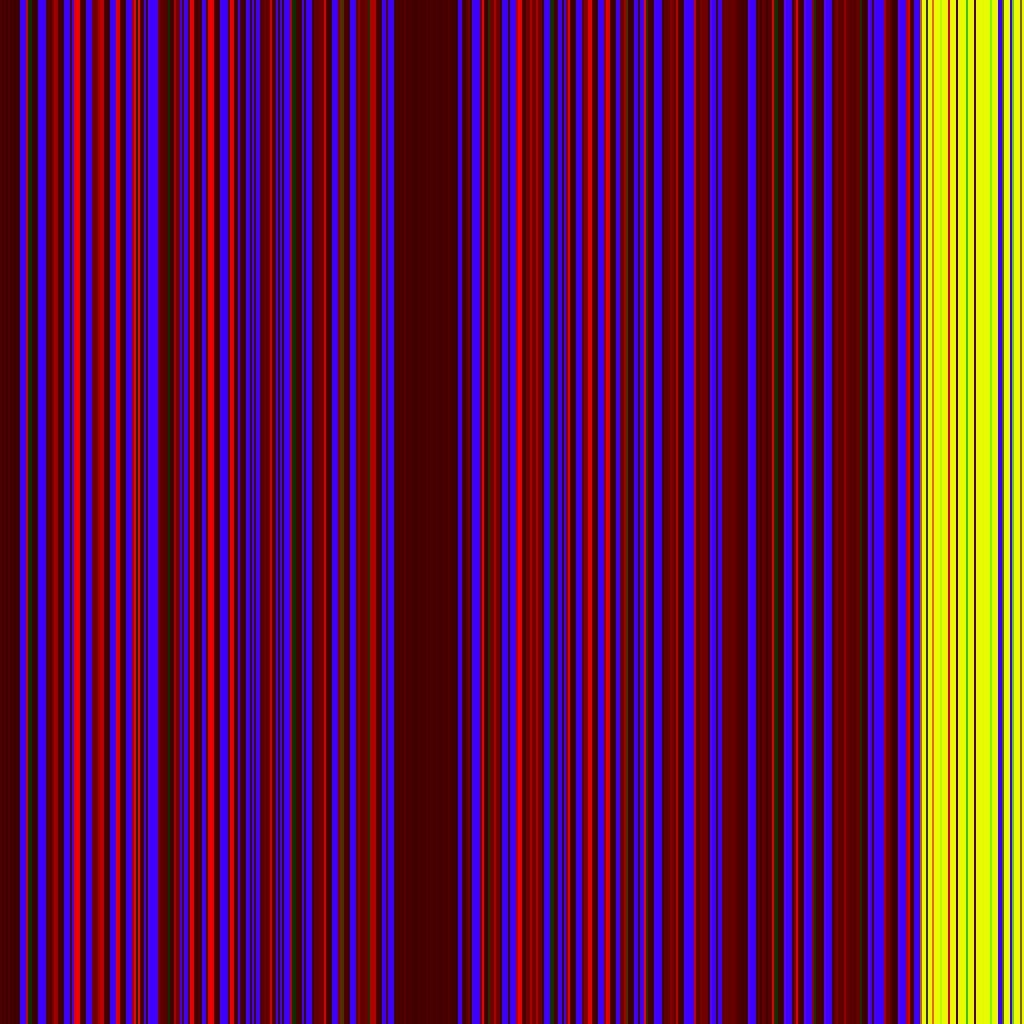}};
    \node[anchor=north west, inner sep=0] at (1.325, 0)
        {\includegraphics[width=2.55cm, height=1.275cm]{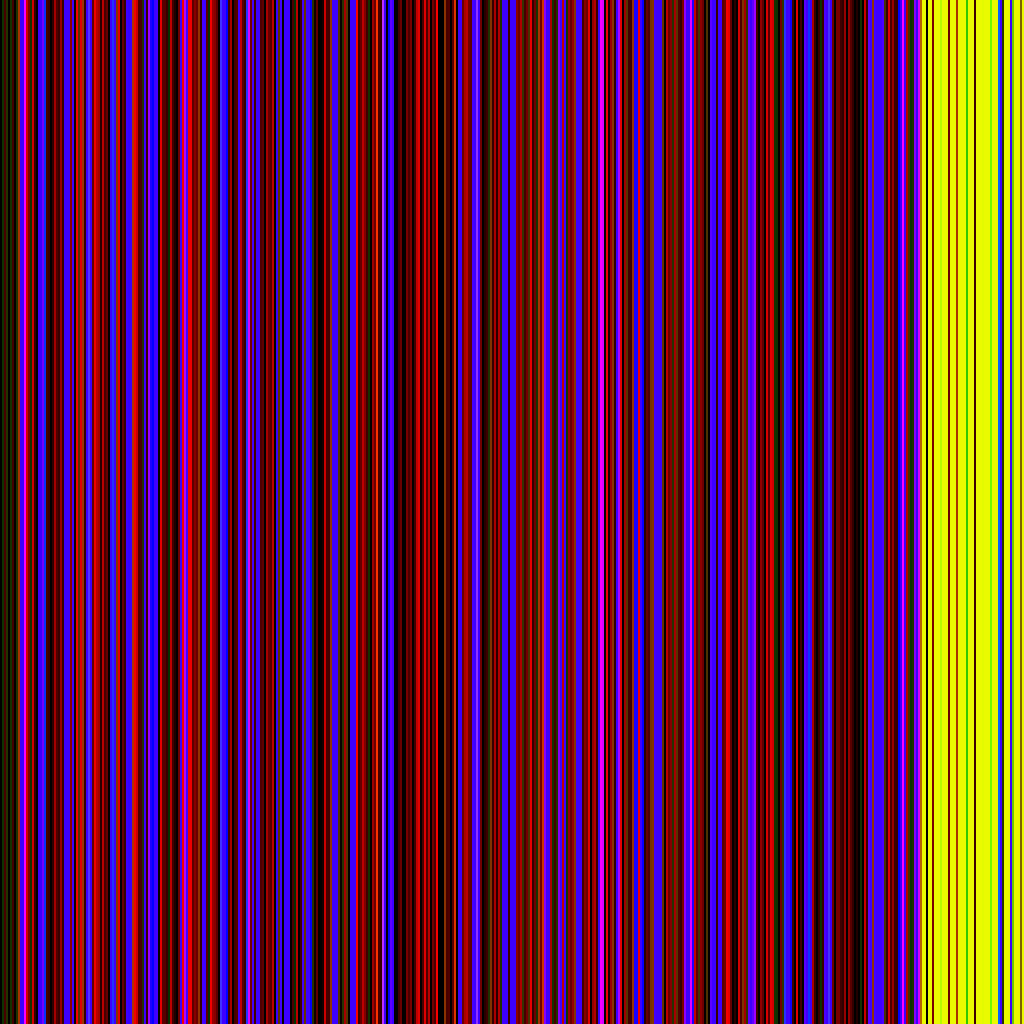}};
    \node[anchor=north west, inner sep=0] at (0, -1.45) {\footnotesize GPT4o};
    \node[anchor=north west, inner sep=2, text=white] at (0, 0) {\fontsize{9pt}{9pt}\selectfont R};
    \node[anchor=north west, inner sep=2, text=white] at (0, -0.70) {\parbox{1.25cm}{\fontsize{9pt}{9pt}\selectfont PK}};
    \node[anchor=north west, inner sep=2, text=white] at (1.325, 0) {\fontsize{9pt}{9pt}\selectfont P};

    \draw[red, very thick] (1.1, 0.05) rectangle (1.25, -1.35);
    \draw[red, very thick] (1.35+2*1.1, 0.05) rectangle (1.35+2*1.25, -1.35);
\end{scope}

\end{tikzpicture}
\caption{Comparison of Raw Content (\textbf{R}), Prior Knowledge (\textbf{PK}), and Phantom (\textbf{P}). 
Significant differences are \textcolor{red}{\fbox{marked}}.
}
\label{fig:calibrated_images}
\vspace{-3em}
\end{figure}

One of the primary challenges in fuzzing is the efficient specification of test cases, as a large number of meaningless or redundant test cases can significantly increase testing overhead. 
Phantom enables the synthesis of samples that are quantitatively close to user-specified test cases.
As shown in Fig.~\ref{fig:calibrated_images}, the final synthesized samples exhibit patterns not present in the originally generated content, demonstrating Phantom's ability to efficiently generate fuzzing test cases.


\begin{figure*}[t]
\centering
\setlength{\abovecaptionskip}{2pt}
\setlength{\belowcaptionskip}{0pt}

\newcommand{\InterCaseGap}{0.45em}

\newcommand{\CaseTag}[1]{%
  \begingroup
  \setlength{\fboxsep}{1.2pt}%
  \colorbox{gray!12}{\textcolor{red}{\strut \footnotesize\bfseries #1}}%
  \endgroup
}

\begin{overpic}[height=0.14\textheight,width=\linewidth,trim=0 2 0 1,clip]{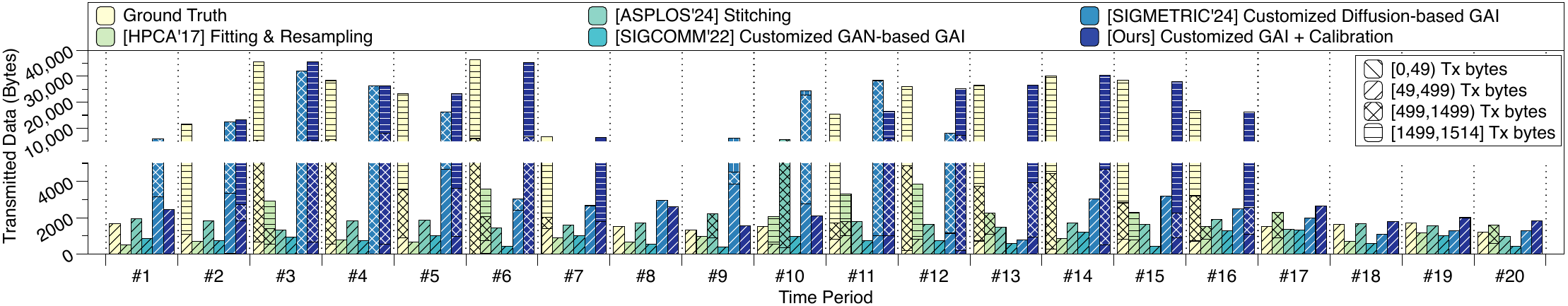}
  \put(6,12.5){\CaseTag{Case 1: H2D}}
\end{overpic}

\vspace{\InterCaseGap}

\begin{overpic}[height=0.12\textheight,width=\linewidth,trim=0 2 0 1,clip]{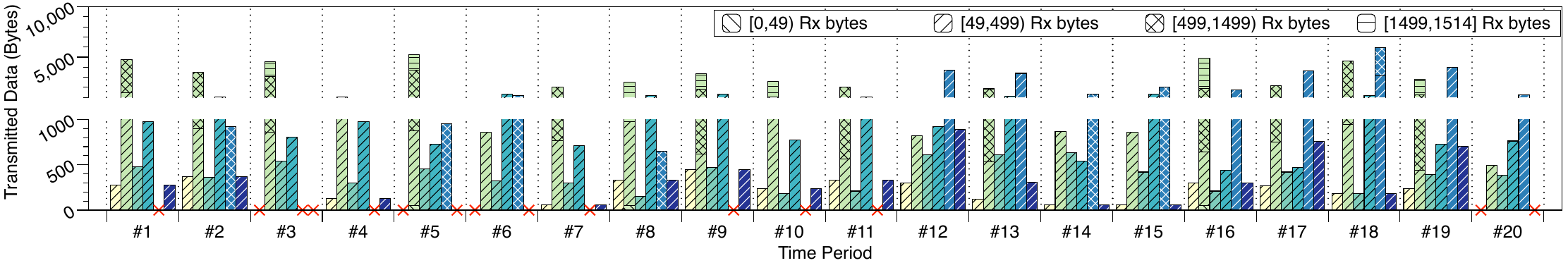}
  \put(6,12.5){\CaseTag{Case 2: D2H}}
\end{overpic}

\caption{Traffic Trends and Breakdown in Synthesized Trace Using Phantom and Four Existing Methods. The trace synthesized by Phantom closely aligns with the ground truth, significantly outperforming the four existing methods.}
\label{fig:previous-methods}
\vspace{-1.25em}
\end{figure*}

\begin{figure*}[t!]
\centering
\setlength{\abovecaptionskip}{1pt}
\setlength{\belowcaptionskip}{-12pt}
\begin{minipage}[t]{0.31\linewidth}
    \centering
    \includegraphics[width=\linewidth,height=0.22\textheight,keepaspectratio]{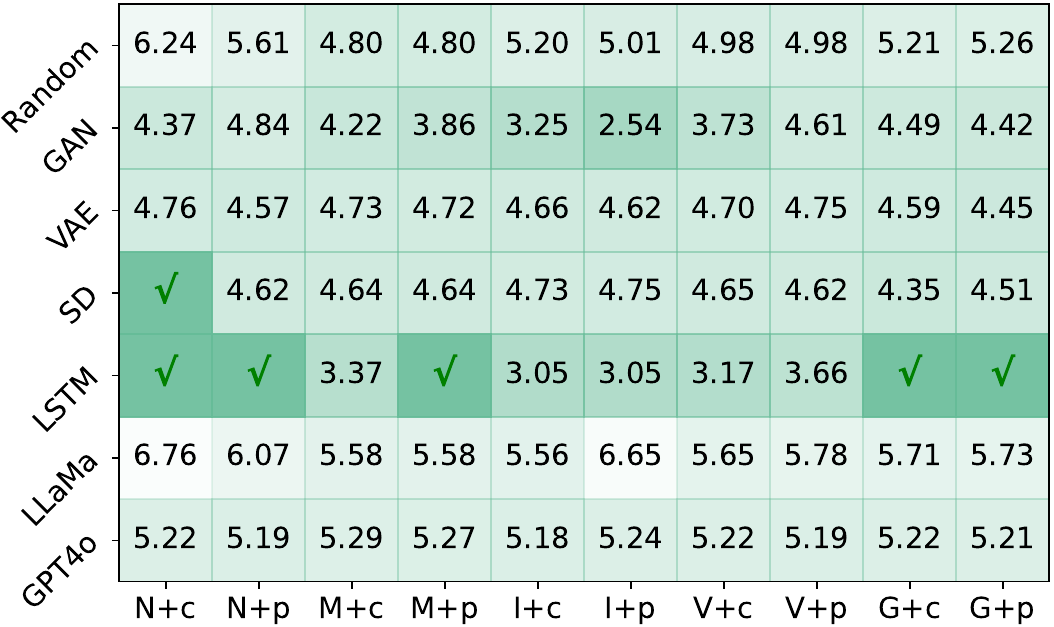}   
    \caption{Heatmap (logarithm) of Transmission Traffic Errors ($\mathit{TE}$). 
    }
    \vspace{-0.2em}
    \label{fig:transmission_traffic_error}
\end{minipage}\hfill
\begin{minipage}[t]{0.31\linewidth}
    \centering
    \includegraphics[width=\linewidth,height=0.2\textheight,keepaspectratio]{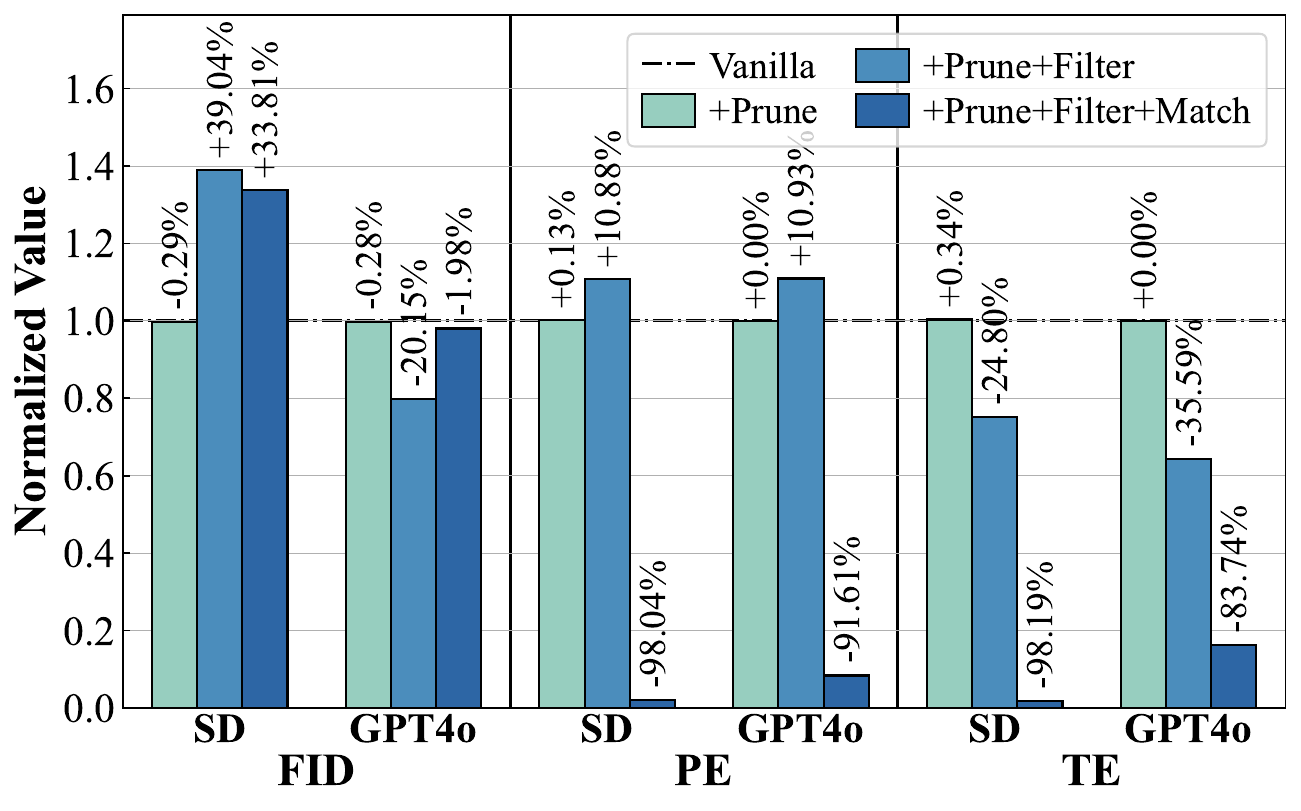}
    \caption{Improvements Attained from Phantom's Components. 
    }
    \vspace{-0.2em}
    \label{fig:ablation_study}
\end{minipage}\hfill
\begin{minipage}[t]{0.31\linewidth}
    \centering
    \includegraphics[width=\linewidth,height=0.15\textheight,keepaspectratio]{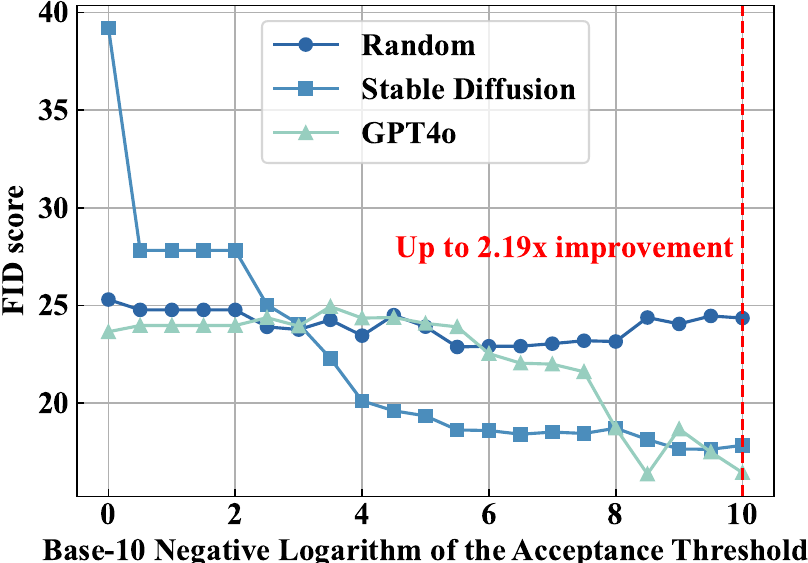}   
    \caption{Generation Quality under Different Thresholds. 
    }
    \vspace{-0.2em}
    \label{fig:convergence_study}
\end{minipage}
\end{figure*}

\subsubsection{Synthesizing Adversarial Traces}

\begin{figure}[H]
\centering
    \setlength{\abovecaptionskip}{0.35cm}
    \setlength{\belowcaptionskip}{-0.6cm}
\includegraphics[width=0.97\linewidth]{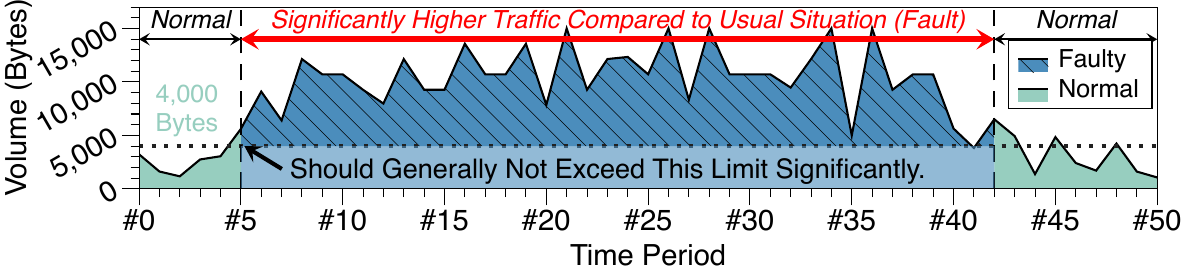}
\vspace{-1.35em}
\caption{A Faulty Trace Example. 
}
\label{fig:fault_inject}
\end{figure}

Phantom can deliberately synthesize adversarial traces by supplying an exemplar with the desired anomaly and lowering the threshold $\lambda$. 
As $\lambda$ decreases, the generated trace converges on the exemplar, progressively embedding forbidden patterns. 
At the driver level, inserting brief faulted fragments into an otherwise normal run lets Phantom emulate software-detectable errors.
Because the captured traces are purely software-level, faults that hinge on hardware corruption or DMA/IOMMU traffic cannot be reproduced unless hardware-level traces are provided. In a proof-of-concept (Fig. ~\ref{fig:fault_inject}), Phantom created a trace whose in-flight traffic briefly exceeded the 4 KB limit, illustrating its utility for counterfactual fault-injection when paired with a hardware/software replay tool.

\subsection{Comparison with Traditional Trace Synthesis Methods}

As previously mentioned, there is a current lack of research specifically focused on PCIe trace generation. We referenced trace generation approaches from other domains and adapted them for PCIe TLP trace synthesis, making minimal modifications.

The methods selected for comparison include: \textbf{1).} Fitting and Resampling (HPCA'17 \cite{maeda2017fast}) \textbf{2).} Stitching (Thesios \cite{10.1145/3620666.3651337}) \textbf{3).} GAN-based Generative AI (NetShare \cite{10.1145/3544216.3544251}) \textbf{4).} Diffusion-based Generative AI (NetDiffusion \cite{jiang2024netdiffusion}). Regarding the experimental setup, all synthesizers utilized the same dataset for fitting, training, or fine-tuning. Specifically, the fitting model employed a Hidden Markov Model, the customized GAN adopted the same architecture as NetShare, and the customized Diffusion model maintained the architecture consistent with NetDiffusion. For Phantom, the hyperparameter $\lambda$ was set to $1 \times 10^{-2}$, while other parameters remained consistent with prior experiments.

As illustrated in Fig.~\ref{fig:previous-methods}, only the traces synthesized by Phantom closely matched the ground truth in both traffic trends and data packet proportions across all time points. In contrast, the other synthesizers demonstrated varying degrees of content distortion. We attribute this discrepancy to the inability of existing methods to effectively map the constraints inherent to the PCIe environment into the trace generation process.

\subsection{Measuring Trace Synthesizing Capability}

\subsubsection{Experiment on Task-Specific Metrics}
We test Phantom's ability to correct AI hallucinations by evaluating its performance over using backbone generators without calibration. Several neural network generators and a baseline random generator with an acceptance threshold of $1\times10^{-8}$ are evaluated. As shown in Tab.~\ref{tab:transmisson_package_error}, $\mathit{PE}$ typically improves by double digits. For VAE and LSTM, Phantom often eliminates $\mathit{PE}$, effectively removing all detected hallucinations by using real data to correct singularities. Fine-tuned feature extractors perform better, emphasizing the value of using similar ground truth data for calibration.

Regarding $\mathit{TE}$, as shown in Fig.~\ref{fig:transmission_traffic_error}, the trend mirrors the $\mathit{PE}$ results. After calibration, Stable Diffusion and LSTM have the lowest $\mathit{TE}$, consistent with their lowest Package Error. Image generation models perform better due to effective capture for TLP patterns.

\subsubsection{Ablation Study}

To assess the impact of Phantom's components on generation quality, we conduct tests on several cases under different setups. We select feature extractor configurations that best optimize $\mathit{TE}$. As shown in Fig.~\ref{fig:ablation_study}, Phantom's performance on FID appears suboptimal. This is because the FID metric is not designed for TLP trace generation tasks, and our configurations are optimized for minimizing $\mathit{TE}$ rather than FID. Notably, simply adding the filter for calibration does not always improve performance, as the prior knowledge samples used might differ too much from the generated content. However, with embedding matching, the similarity between the two increases significantly, making calibration more effective and resulting in substantial improvement.

\subsubsection{Generation Quality Convergence}

As shown in Fig.~\ref{fig:convergence_study}, except for the worst-case random generation, both Stable Diffusion and GPT4o exhibit a decreasing trend in FID after Phantom calibration. This demonstrates Phantom's ability to align generated content more closely with real TLP trace with user requirements.

\section{Discussions}
\noindent \textbf{Validation on simulator or different device.} Replay-based validation would strengthen the evaluation. Current hardware lacks native PCIe tracing support on both the root complex and endpoint, and FPGA prototyping still falls short in trace fidelity and throughput. As a practical alternative \cite{erbsen2021integration,song2019periscope}, we currently use software-level tracing via MMIO/PIO, which is broadly applicable to PCIe devices and suitable for emulation environments like VirtIO, though it may not capture all hardware-specific timing artifacts.

\noindent \textbf{Future Work.}
Phantom’s design is adaptable to other fields, and similar generation-and-calibration methods can be applied to other buses or networks. We plan to explore adapting Phantom to Kubernetes and SCSI hard drives. To enable more targeted performance analysis, we also plan to incorporate additional parameters such as frequency and latency.

\section{Conclusion}
We introduce Phantom, a novel framework that makes generative AI practical for device simulation by solving the problem of AI hallucination during PCIe TLP trace synthesis. The core of Phantom's methodology is a two-stage process: a generative backbone creates trace data, which is then refined by a domain-specific post-processing filter. This filter enforces strict protocol compliance, guaranteeing that the final outputs are functionally valid. Through an evaluation on real-world NIC traces, we show that Phantom significantly mitigates hallucinations, outperforming backbone-only methods  by up to 1000$\times$ on task-specific metrics and up to 2.19$\times$ on Fréchet Inception Distance. The resulting traces are thus both statistically faithful and directly usable in high-fidelity simulations.

\begin{acks}
This project is supported by the National Key R\&D Program of China (2022YFB4402102), Ministry of Industry and Information Technology 2024 Cloud Operating System Project, the National NSF of China (No. 62572307, No.62402311), Natural Science Foundation of Shanghai (No.24ZR1433700), Key Research and Development Program of Shanghai (25LN3201200), and Shanghai Key Laboratory of Scalable Computing and Systems. The corresponding authors are Fangxin Liu and Zhengwei Qi.
\end{acks}

\bibliographystyle{ACM-Reference-Format}
\bibliography{phantom}










\end{document}